\PassOptionsToPackage{table}{xcolor}
\documentclass[11pt]{article}

\usepackage{acl}
\usepackage{times}
\usepackage{latexsym}
\usepackage[T1]{fontenc}
\usepackage[utf8]{inputenc}
\usepackage{textcomp}
\usepackage{microtype}
\usepackage{inconsolata}
\usepackage{graphicx}
\usepackage{amsmath}
\usepackage{amssymb}
\usepackage{booktabs}
\usepackage{multirow}
\usepackage{xcolor}
\usepackage{placeins}
\usepackage{eso-pic}
\usepackage[most]{tcolorbox}
\definecolor{myblue}{RGB}{55,138,221}
\definecolor{tblBlue}{HTML}{E8F1FF}
\definecolor{tblGreen}{HTML}{E7F6EC}
\definecolor{tblHead}{HTML}{223A5E}
\definecolor{preprintGray}{HTML}{59636E}
\newcommand{\preprintmark}{%
  \AddToShipoutPictureFG*{%
    \AtPageUpperLeft{%
      \raisebox{-0.44in}[0pt][0pt]{%
        \makebox[\paperwidth][c]{%
          \sffamily\fontsize{9}{10}\selectfont\color{preprintGray}%
          \textit{Preprint}\hspace{0.55em}{\scriptsize\textbullet}\hspace{0.55em}2026-11-26%
        }%
      }%
    }%
  }%
}
\newcommand{\meanstd}[2]{\shortstack{#1\\[-2pt]{\scriptsize$\pm #2$}}}
\newcommand{\bestmeanstdup}[3]{\shortstack{\textbf{#1}\raisebox{0.6ex}{\textcolor{myblue}{\scriptsize$\uparrow #3$}}\\[-2pt]{\scriptsize$\pm #2$}}}
\newtcolorbox{findingbox}[1]{
  colback=tblBlue!55,
  colframe=tblHead,
  boxrule=0.5pt,
  arc=1mm,
  left=4pt,
  right=4pt,
  top=3pt,
  bottom=3pt,
  before skip=5pt,
  after skip=5pt,
  fonttitle=\bfseries\small,
  coltitle=white,
  colbacktitle=tblHead,
  title={#1}
}

\title{More Informed Teachers, Less Reliable Supervision:\\
An Empirical Analysis of Privileged On-Policy Distillation\\
for Retrieval-Augmented Generation}

\author{
\textbf{Shaoping Huang\textsuperscript{1,2}},
\textbf{Chenhui Hu\textsuperscript{2,3}},
\textbf{Zhuoran Jin\textsuperscript{2,3}},
\textbf{Jiachun Li\textsuperscript{2,3}} \\
\textbf{Chenhao Yuan\textsuperscript{2,3}},
\textbf{Tianyi Men\textsuperscript{2,3}},
\textbf{Shizhu He\textsuperscript{2,3}},
\textbf{Jun Zhao\textsuperscript{2,3}} \\
\textsuperscript{1}School of Advanced Interdisciplinary Sciences,
University of Chinese Academy of Sciences \\
\textsuperscript{2}The Key Laboratory of Cognition and Decision Intelligence for Complex Systems, \\
Institute of Automation, Chinese Academy of Sciences \\
\textsuperscript{3}School of Artificial Intelligence,
University of Chinese Academy of Sciences \\
\texttt{huangshaoping2025@ia.ac.cn}\quad
\texttt{\{zhuoran.jin,shizhu.he,jzhao\}@nlpr.ia.ac.cn}
}

\hypersetup{
  pdftitle={More Informed Teachers, Less Reliable Supervision: An Empirical Analysis of Privileged On-Policy Distillation for Retrieval-Augmented Generation},
  pdfauthor={Shaoping Huang, Chenhui Hu, Zhuoran Jin, Jiachun Li, Chenhao Yuan, Tianyi Men, Shizhu He, Jun Zhao}
}

\begin{document}
\preprintmark
\maketitle

\begin{abstract}
On-policy distillation (OPD) trains a student by asking a teacher to score the
student's own generated response. Privileged OPD additionally gives the teacher
gold answers or supporting evidence, with the natural expectation that more
information yields better supervision. Unlike supervised fine-tuning, however,
OPD updates the student's sampled tokens rather than the teacher's answer. We
uncover a counterintuitive phenomenon: \textbf{\emph{a more informed teacher can
produce less useful OPD supervision}}. Across three complete
development sets and three seeds, \textbf{adding gold evidence does not
reliably improve OPD}; on HotpotQA, it reduces F1 by $2.05$ under otherwise
matched conditions. We explain this reversal through \emph{reachability},
\emph{projection}, and \emph{compatibility}. The more informed teacher selects
correct responses more often yet can \textbf{assign incompatible credit to
their answer tokens}. Matching the answer sequence scored by OPD and supervised
fine-tuning \textbf{contracts $93.8$--$99.3\%$ of their layer-wise
gradient-coherence gap} across four datasets. Matched interventions show that
\textbf{reward rescaling alone is insufficient}. Answer-space projection and
answer-aware selection help only when
they target the active bottleneck. Overall, the \textbf{supervised sequence}
links teacher knowledge to student learning and determines whether privilege
yields useful OPD updates.
\end{abstract}

\section{Introduction}
\label{sec:introduction}

\begin{figure}[!t]
\centering
\includegraphics[width=\columnwidth]{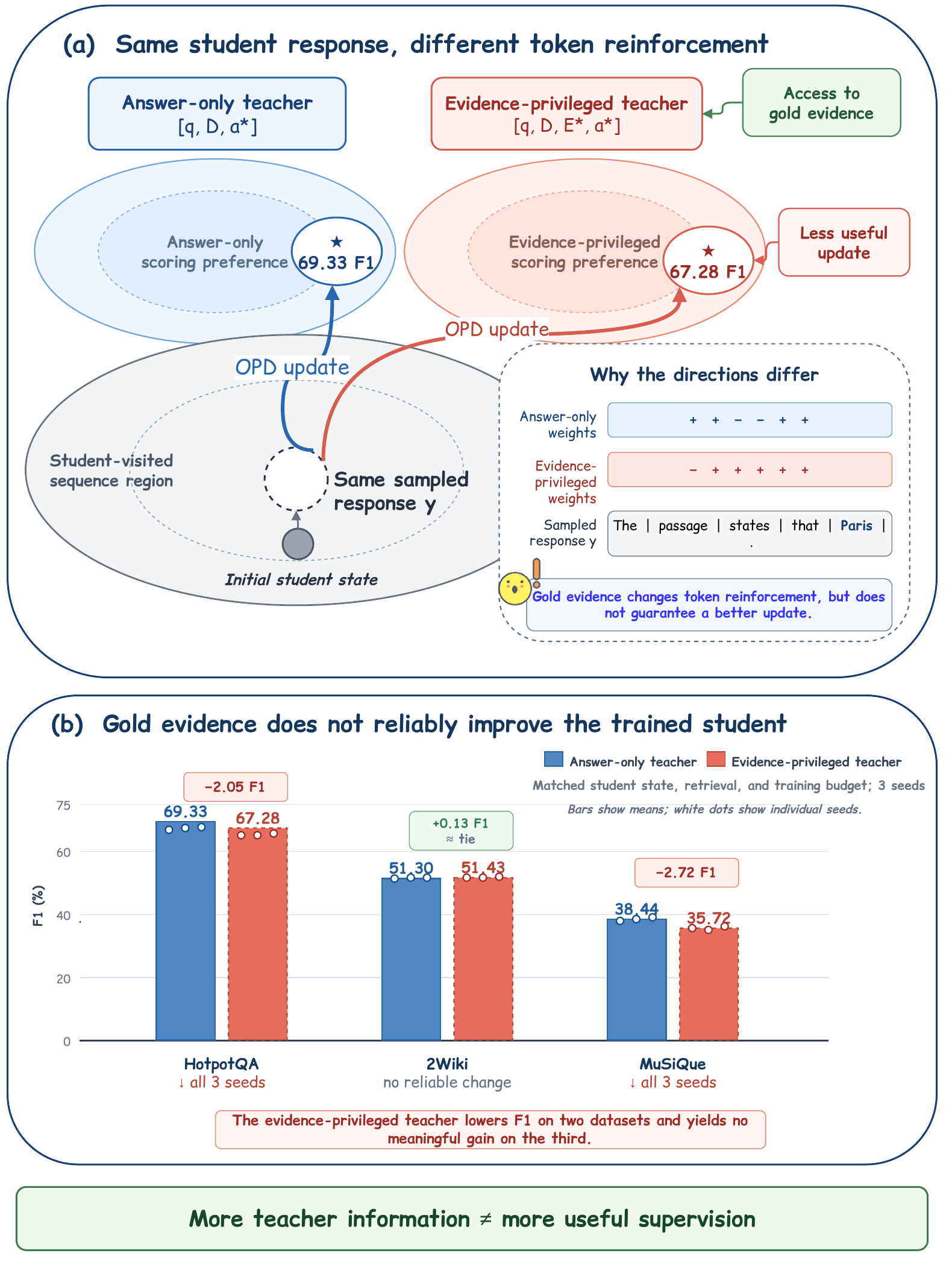}
\caption{More teacher information does not reliably improve OPD supervision.
(a) Two teachers score the same response differently. (b) Evidence
privilege changes F1 by $-2.05$, $+0.13$, and $-2.72$ on HotpotQA, 2Wiki, and
MuSiQue (three-seed means).}
\label{fig:main-overview}
\end{figure}

Knowledge distillation commonly trains a student to imitate teacher-generated
demonstrations. On-policy distillation (OPD) instead asks the teacher to score
the student's own responses, so supervision follows the states and mistakes
the student actually encounters
\citep{agarwal2024policy,gu2024minillm,ko2024distillm}. Recent work improves
this process through better objectives and token selection
\citep{zhang2026full,pan2026rlcsd,li2026rethinking,zhu2026many,
wang2026demystifying,li2026filter}; privileged distillation gives teachers
training-only answers, rationales, or context
\citep{zhao2026self,penaloza2026privileged,ye2026policy}. Combining the two
appears natural: a better-informed teacher should better evaluate the student's
behavior. Yet OPD does not train on the teacher's answer; it trains on tokens
already generated by the student. Whether privileged knowledge survives this
conversion remains unclear. \textbf{Does more teacher information necessarily
yield better OPD supervision?}

Multi-hop retrieval-augmented generation (RAG) provides a controlled setting
for this question: each example pairs a short answer with gold supporting
documents \citep{lewis2020retrieval,yang2018hotpotqa,ho2020constructing,
trivedi2022musique}. We compare answer-only and evidence-and-answer teachers
while holding the student model, retrieved documents, training examples, and
optimization budget constant. Contrary to expectation
(Figure~\ref{fig:main-overview}), gold evidence changes mean F1 by $-2.05$ on
HotpotQA, $-2.72$ on MuSiQue, and only $+0.13$ on 2Wiki. Answer-only OPD also
wins all three complete HotpotQA development runs
(Table~\ref{tab:teacher-privilege}). Thus, \textbf{gold evidence makes the
teacher more informed, but does not reliably make OPD supervision more
effective.}

A conditional-distribution shift explains why two teachers prefer different
tokens, but not why the more informed teacher produces a weaker student. The
distinction becomes clear when compared with supervised fine-tuning (SFT).
SFT maximizes the likelihood of a known answer, whereas OPD assigns signed
teacher weights throughout a sampled free-form response. Such a response may
contain the correct answer together with an explanation, evidence restatement,
or alternative wording. The teacher can therefore recognize the response as
correct while assigning unfavorable weight to the tokens that express its
answer.
\textbf{Where is useful supervision lost between teacher judgment and the
student update?}

We analyze three requirements: \emph{reachability}, the rollout contains a
usable answer; \emph{projection}, that answer is recovered as the task-relevant
candidate; and \emph{compatibility}, the token weights support that candidate.
These requirements are sequential: OPD cannot train an answer that was not
generated, cannot target one that was not recovered, and cannot improve one
that receives an opposing update. Matched traces expose the final failure: the
evidence-and-answer teacher selects correct responses more often than random
($60.4\%$ vs. $56.1\%$), yet $57.6\%$ of its selected correct answer spans
receive negative net weight. Better judgment therefore does not guarantee
compatible supervision.

We connect this compatibility failure to SFT by controlling the scored
sequence. When SFT and OPD score the same normalized gold answer, their
layer-wise gradient-coherence gap contracts by $93.8$--$99.3\%$ across four
datasets (Figure~\ref{fig:decision-aligned-credit}). Shortening a rollout to the
same token count explains much of this scalar contraction, but its update
direction remains poorly aligned with SFT; shared answer content is required
for directional alignment. Thus SFT directly trains the answer, whereas OPD
must first satisfy reachability, projection, and compatibility on a sampled
response.

Finally, three interventions test this account. Reward rescaling does not
remove the teacher ordering; projection is strongly dataset dependent; and
compatibility-aware token selection beats a rate-matched random gate on three
of four full evaluation sets. Privileged information is useful only when it
reaches the sequence and token updates that define the task.

\paragraph{Contributions.} Our analysis proceeds from phenomenon to mechanism
and intervention: \textbf{(i)} matched three-seed evaluations on three complete
development sets establish that richer teacher information does not reliably
improve OPD; \textbf{(ii)} reachability, projection, compatibility, and
matched-sequence analysis connect this failure to the SFT--OPD gap across four
datasets; and \textbf{(iii)} targeted controls distinguish reward magnitude,
sequence projection, and answer-aware token selection as separate bottlenecks.

\begin{figure*}[t]
\centering
\includegraphics[width=\textwidth]{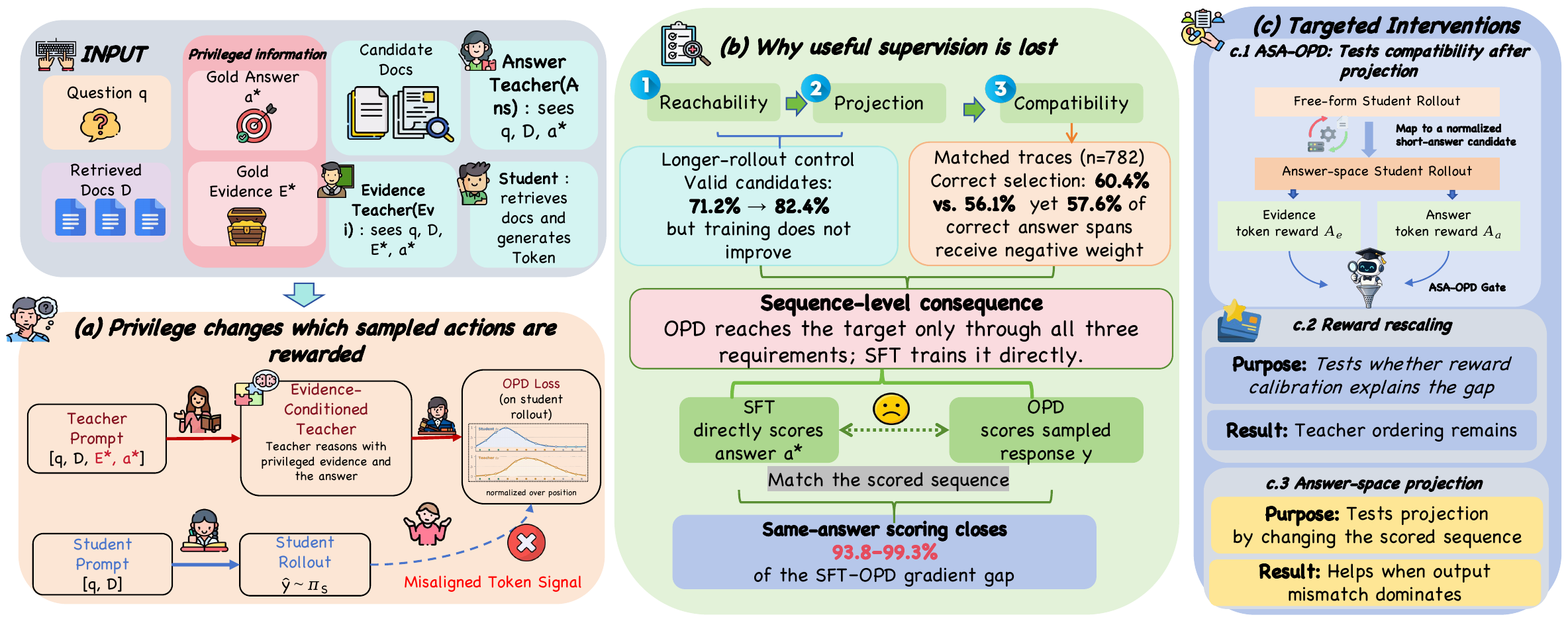}
\caption{Overview of our analysis. \textbf{(a)} Additional teacher information
can weaken OPD under the same student state. \textbf{(b)} We trace useful
supervision through reachability, projection, and compatibility, and connect
OPD to SFT by matching the scored sequence. \textbf{(c)} Targeted interventions
isolate the corresponding bottlenecks.}
\label{fig:three-stage}
\end{figure*}

\section{Preliminaries: RAG On-Policy Distillation}
\label{sec:setup}

\subsection{Privileged Teachers in Multi-Hop RAG}

Let $q$ be a question, $D=\{d_1,\ldots,d_k\}$ the documents selected by a
retriever or reranker, and $a$ the reference answer. The answer generator
receives the state $x=(q,D)$ and produces a short answer. We study three
training-only privileges $z$: the gold answer (\emph{answer-only teacher}), all
annotated supporting evidence plus the same answer (\emph{evidence-and-answer teacher}), and
only the supports already present in $D$ plus the answer
(\emph{visible-evidence teacher}). These teachers differ only in privileged
context; the student always acts from the same retrieved state.
The prompt analysis additionally removes the answer, shortens the evidence prompt
while preserving all annotated supports, or exposes only supports visible in
$D$. These variants are prompt controls rather than additional teacher
definitions.

\subsection{On-Policy Distillation from Student Rollouts}

The student samples a continuation $\hat y\sim\pi_S(\cdot\mid x)$. At token
position $t$, the teacher scores the student-sampled token under privilege $z$:
\[
\resizebox{0.98\columnwidth}{!}{$\displaystyle
A_t(z)=\log\pi_T(\hat y_t\mid x,z,\hat y_{<t})-\log\pi_S(\hat y_t\mid x,\hat y_{<t}).
$}
\]
For any scored sequence $u$ and detached token weight $w$, define the common
weighted-likelihood objective
\[
\resizebox{0.98\columnwidth}{!}{$\displaystyle
\mathcal L(u,w)=-\sum_t \mathrm{sg}(w_t)
\log\pi_S(u_t\mid x,u_{<t}).
$}
\]
Here $\mathrm{sg}$ denotes stop-gradient. Direct answer supervision uses
$(u,w)=(a,\mathbf 1)$; sampled-token OPD uses
$(u,w)=(\hat y,\operatorname{clip}_{[-c,c]}A(z))$. Thus, SFT supplies both the
target answer and positive weight, whereas OPD must obtain a useful answer from
the rollout and compatible token weights from the teacher. Without clipping, the OPD
expectation has the gradient of a reverse-KL objective on student-visited
prefixes \citep{gu2024minillm,agarwal2024policy}. Our analysis is scoped to
this sampled-token setting rather than all sequence-level or full-vocabulary
distillation objectives.

\begin{table}[t]
\centering
\caption{HotpotQA teacher-privilege comparison. Panel (a) uses seed 42;
panel (b) reports matched three-seed means and standard deviations. Blue
superscripts denote Answer only minus Evidence + answer.}
\label{tab:teacher-privilege}
\scriptsize
\textbf{\textit{(a) Privilege sweep: diagnostic set, seed 42}}\par\vspace{3pt}
\begin{tabular*}{\columnwidth}{@{\extracolsep{\fill}}lcc}
\toprule
\textbf{Teacher input} & \textbf{F1} & \textbf{EM} \\
\midrule
None & 44.87 & 30.20 \\
Gold evidence + answer & 61.33 & 47.20 \\
Visible evidence + answer & 50.26 & 37.20 \\
Answer only & \textbf{62.82} & \textbf{49.80} \\
\bottomrule
\end{tabular*}

\vspace{6pt}
\textbf{\textit{(b) Principal comparison across three seeds}}\par\vspace{3pt}
\renewcommand{\arraystretch}{1.10}
\begin{tabular*}{\columnwidth}{@{\extracolsep{\fill}}llcc}
\toprule
\textbf{Evaluation} & \textbf{Metric} & \textbf{Evidence + answer} & \textbf{Answer only} \\
\midrule
\multirow{2}{*}{Diagnostic} & F1 & \meanstd{61.26}{0.37} & \bestmeanstdup{63.18}{0.28}{1.92} \\
& EM & \meanstd{47.47}{0.52} & \bestmeanstdup{50.07}{0.25}{2.60} \\
\addlinespace[1pt]
\multirow{2}{*}{Full dev.} & F1 & \meanstd{67.28}{0.29} & \bestmeanstdup{69.33}{0.33}{2.05} \\
& EM & \meanstd{53.13}{0.57} & \bestmeanstdup{55.25}{0.66}{2.12} \\
\bottomrule
\end{tabular*}
\end{table}

\section{More Informed Teachers Do Not Reliably Improve OPD}
\label{sec:observation}

We first test the assumption that additional privileged information necessarily
produces better on-policy supervision. The comparison holds the student and
its retrieved context constant, so any difference comes from how the teacher
scores the same student-generated sequences.

\paragraph{Evaluation protocol.}

We evaluate HotpotQA, 2WikiMultiHopQA, MuSiQue, and QASC
\citep{yang2018hotpotqa,ho2020constructing,trivedi2022musique,khot2020qasc}.
The answer generator is Qwen3-14B with LoRA adaptation
\citep{yang2025qwen3,hu2022lora}. Unless noted otherwise, it samples at
temperature $1.0$, top-$p=0.95$, top-$k=20$, with at most 32 new tokens.
We report normalized F1 and exact match (EM). Experiments use either the full
evaluation set or a diagnostic set, as labeled. Repeated baseline names in
separate ablation blocks denote
independently trained runs; each intervention is compared only with the
baseline from the same block. Prompts and optimization details are given in
Appendix~\ref{app:impl}.

\subsection{Additional Gold Evidence Can Reduce OPD Performance}

\paragraph{Setup.}
We first vary teacher privilege on the HotpotQA diagnostic set, then repeat the
principal evidence-and-answer versus answer-only comparison across three seeds
on both the diagnostic and full development sets
(Table~\ref{tab:teacher-privilege}). Every comparison keeps the student model,
retrieved documents, training pool, and update budget unchanged; retrieval
quality therefore cannot explain the difference.

\paragraph{Results.}
Performance is not monotonic in privileged information. After applying the
same format-robust answer parser to every condition, answer-only reaches
$63.18{\pm}0.28\%$ F1 and gold-evidence-plus-answer reaches $61.26{\pm}0.37\%$ on the
diagnostic set. The full-development comparison is not a single-run artifact:
answer-only reaches $69.33{\pm}0.33\%$ versus $67.28{\pm}0.29\%$ for the
evidence-and-answer teacher, and wins in every seed. The effect is therefore smaller on the full
set but remains consistent under the same student state. \textbf{Additional gold
evidence therefore does not reliably improve OPD, even when the student state
and training budget are unchanged.}

\begin{findingbox}{Finding 1: More Information Can Weaken OPD}
Additional teacher information can weaken OPD even when the student and its
retrieved state remain unchanged.
\end{findingbox}

\section{Why Better Teacher Judgment Does Not Guarantee Better Student Updates}
\label{sec:mechanism}

The matched comparison establishes the reversal but not where supervision is
lost. We test \emph{reachability} (whether the rollout contains a usable
answer), \emph{projection} (whether that answer is recovered), and
\emph{compatibility} (whether its token weights support the answer). We then
connect these requirements to SFT by matching the scored sequence.

\begin{figure}[t]
\centering
\includegraphics[width=\columnwidth]{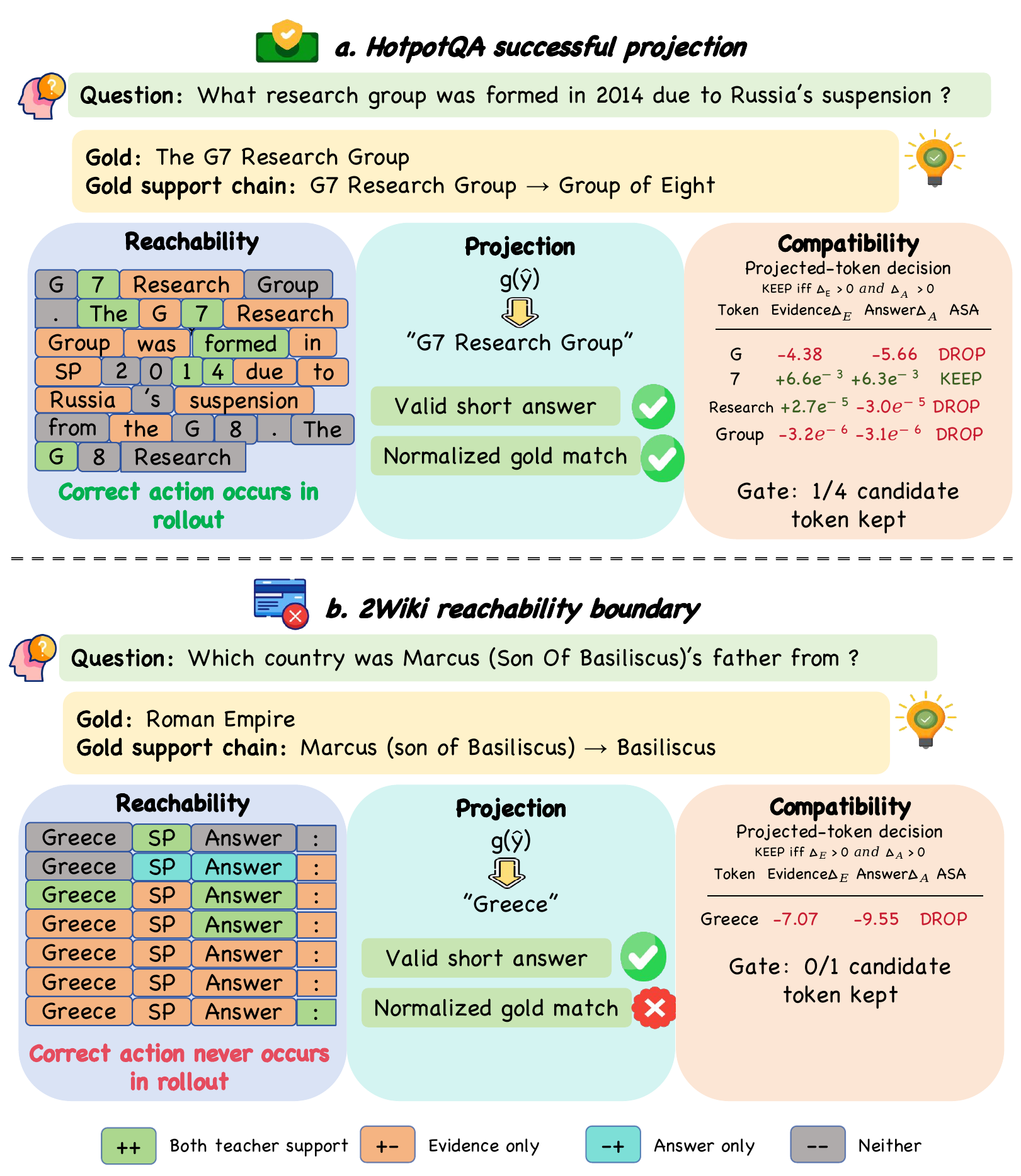}
\caption{Two training traces instantiate distinct failure modes in privileged OPD.
\textbf{(a)} The HotpotQA rollout contains the correct answer and projection
maps it to a normalized candidate, but the two teachers jointly support only one of its four tokens.
\textbf{(b)} None of the sampled 2Wiki rollouts contains the reference answer;
projection therefore returns an incorrect candidate and no token is retained.
The quantitative analyses below test how often these two boundaries occur.}
\label{fig:case-mechanism}
\end{figure}

\subsection{Reachability and Projection: Recovering a Usable Answer}
\label{sec:reachability}
\label{sec:projection}

Sampled-token OPD can reinforce only tokens generated by the student. We
therefore measure whether at least one of eight rollouts contains the
normalized gold answer. On the HotpotQA full evaluation set, increasing the
maximum rollout length from 32 to 128 tokens raises answer occurrence from
$69.3\%$ to $81.1\%$ (Appendix Figure~\ref{fig:rollout-length-full-eval}). The increase
shows that limited sampling excludes a substantial fraction of correct answers,
but occurrence alone does not determine whether the answer can be
used for short-answer training. A rollout may contain the answer inside an explanation without yielding a
usable short answer. We apply the answer extractor and separately measure
whether it returns a valid short candidate and whether that candidate exactly
matches the normalized gold answer. From 32 to 128 tokens, valid extraction
rises from $71.4\%$ to $82.6\%$, while exact extraction rises from $41.1\%$
to $47.3\%$ (Appendix Figure~\ref{fig:rollout-length-full-eval}).
\textbf{Longer rollouts improve both reachability and projection coverage, but
more than one third of examples still lack an exact extracted answer.}

The limitation is dataset-dependent. On the QASC full evaluation set, matched
three-seed evaluation increases answer-only OPD from
$17.41{\pm}2.04\%$ F1 at 32 tokens to $41.14{\pm}1.15\%$ at 128 tokens.
Evidence-and-answer OPD increases by a similar amount, from
$21.17{\pm}0.91\%$ to $45.20{\pm}0.88\%$. Longer decoding therefore explains
QASC's low absolute scores under the 32-token limit, but not the teacher
ordering: evidence-and-answer remains ahead by $3.76$ F1 at 32 tokens and
$4.07$ F1 at 128 tokens. We analyze QASC as a closed-answer boundary case in
Appendix~\ref{app:impl} rather than as evidence for the open-answer reversal.

\subsection{Compatibility Begins with Correct-Response Selection}
\label{sec:compatibility}

We next ask whether the evidence-and-answer teacher can recognize a better
student answer. For each HotpotQA prompt, we sample two student rollouts and
compare selection by the evidence-and-answer teacher with seeded random selection. Both
conditions see the same candidates and make the same number of updates. Across
$782$ exactly matched saved traces, the teacher selects a correct extracted
answer in $60.36\%$ of cases, compared with $56.14\%$ for random selection; the
$4.22$-point difference has a bootstrap 95\% confidence interval of
$[1.79,6.65]$. The privileged teacher therefore contains useful information
about which completed answer is better.

\subsection{Compatibility Fails at Answer-Token Reinforcement}

Better selection does not imply a useful OPD update. We rescore the
teacher-selected traces with the final student checkpoint and decompose the
signed OPD weights into answer and non-answer tokens. Among the $472$
teacher-selected traces with an exactly correct answer, $57.63\%$ assign
negative net weight to the answer span and $78.18\%$ assign negative net
weight to the complete rollout. Moreover, $89.42\%$ of all negative weight
lies outside the answer span. \textbf{The teacher more often chooses a correct
response, while ordinary OPD still suppresses most of those responses; the
surrounding explanation contributes most of the penalty.} This analysis is
a final-checkpoint rescore of real training traces, not a reconstruction of
every historical update.

\begin{figure}[t]
\centering
\includegraphics[width=\columnwidth]{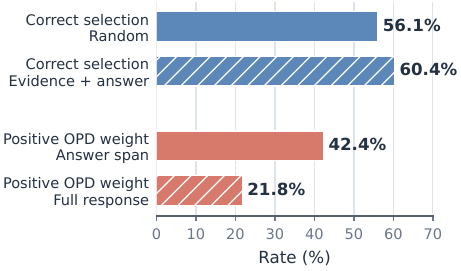}
\caption{Correct-response selection and answer-token reinforcement diverge.
Among $782$ matched traces, the evidence-and-answer teacher selects a correct
response in $60.4\%$ of cases, compared with $56.1\%$ for random selection.
Among the $472$ selected correct responses, only $42.4\%$ of answer spans and
$21.8\%$ of full responses receive positive net OPD weight.}
\label{fig:selection-reinforcement}
\end{figure}

Correct-response selection and positive token supervision are therefore distinct. Appendix
Figure~\ref{fig:pls-ga-mechanism} and Tables~\ref{tab:diagnostic-ci}--
\ref{tab:sequence-compatibility} provide the complementary PLS and
gradient-direction diagnostics.

\begin{findingbox}{Finding 2: Correct Selection Does Not Ensure Compatibility}
A teacher can identify better responses while assigning incompatible credit
to the answer tokens that make them correct.
\end{findingbox}

\subsection{Sequence Length and Content Shape the SFT--OPD Gap}

\begin{figure*}[t]
\centering
\includegraphics[width=0.88\textwidth]{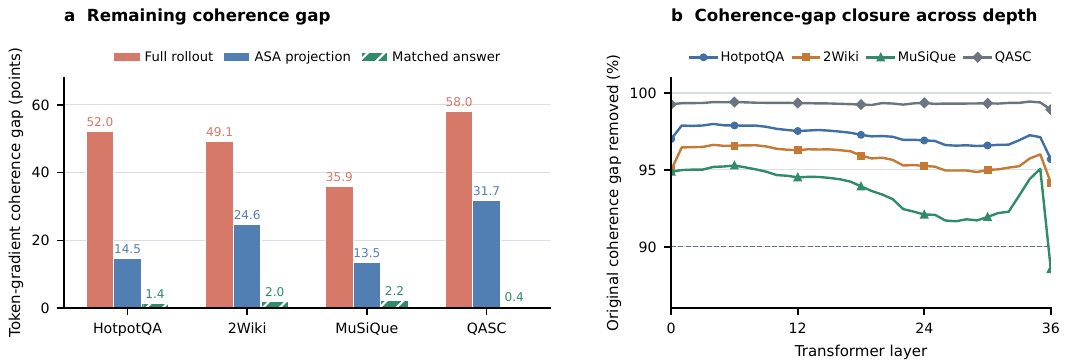}
\caption{Matching the training sequence connects OPD to SFT.
\textbf{(a)} Coherence gaps from answer supervision for a free-form rollout,
its projected answer, and a shared gold answer. \textbf{(b)} Shared-answer
scoring contracts $93.8$--$99.3\%$ of the original gap across 36--37 of 37
transformer layers on each full evaluation set. Appendix
Table~\ref{tab:length-matched-gradient-control} separates this contraction from
the effect of token count.}
\label{fig:decision-aligned-credit}
\end{figure*}

SFT and OPD update the same student from the same retrieved context, but they
differ in two respects. SFT trains the gold answer with uniformly positive
weights, whereas standard OPD trains a free-form student response with signed
teacher weights. We use SFT only to define the desired answer direction, then
control the sequence on which the two objectives are evaluated. If their
gradients remain different on the same answer sequence, the discrepancy must
come from the teacher weights. If the discrepancy becomes small, the choice of
training sequence accounts for most of the original difference. Let
$g_{\mathrm{task}}=\nabla\mathcal L_{\mathrm{SFT}}$ and
$g_{\mathrm{OPD}}=\nabla\mathcal L_{\mathrm{OPD}}$. After one OPD step,
$\theta'=\theta-\eta g_{\mathrm{OPD}}$, the first-order change in the answer
objective is
\[
\mathcal L_{\mathrm{SFT}}(\theta')-\mathcal L_{\mathrm{SFT}}(\theta)
=-\eta\langle g_{\mathrm{task}},g_{\mathrm{OPD}}\rangle+O(\eta^2).
\]
This gives the comparison a direct learning interpretation: an OPD gradient
that points in the SFT direction locally lowers the correct-answer loss,
whereas an opposing gradient raises it. Gradient alignment (GA) reports the
normalized agreement:
\[
\mathrm{GA}=
\cos\!\left(
\nabla\mathcal L_{\mathrm{OPD}},
\nabla\mathcal L_{\mathrm{SFT}}
\right).
\]
Positive and negative GA indicate locally aligned and opposing update
directions, respectively. We compute it on LoRA parameters as an aggregate
diagnostic. In the 2Wiki $D_{\mathrm{all}}$ condition,
answer-only OPD has positive aggregate alignment ($+0.087$), whereas
full-evidence OPD has negative alignment ($-0.031$). The expanded analysis
and support-preserving controls show that visibility alone does not determine
the update direction (Appendix
Tables~\ref{tab:diagnostic-ci}--\ref{tab:strict-matched-state}).

\paragraph{Each objective trains its original sequence.}
Across all four full evaluation sets, layer-wise gradient coherence follows the
same ordering on the sequences actually used for training: direct answer
supervision is highest, ASA-OPD is intermediate, and free-form full-evidence
OPD is lowest (Figure~\ref{fig:decision-aligned-credit}, top). On these original
sequences, answer projection reduces the gap to SFT by $45.4$--$72.1\%$.

\paragraph{All objectives score the same answer sequence.}
We then make all three objectives score the same normalized gold-answer
sequence while retaining their respective token weights. The SFT versus full-evidence gap falls from $35.9$--$58.0$ coherence
points to $0.4$--$2.2$, a reduction of $93.8$--$99.3\%$
(Figure~\ref{fig:decision-aligned-credit}, bottom). The contraction occurs in
at least 36 of 37 layers on every dataset; Appendix
Figure~\ref{fig:layerwise-gradient-coherence} reports the complete per-layer curves.

\paragraph{Interpretation.}
SFT defines the reference direction for the desired answer update; it is not
treated as an objective that OPD must imitate. Coherence alone is sensitive to
sequence length: exact length-matched rollout spans close $55.4$--$82.4\%$ of
the scalar gap on multi-token answers. However, their best mean layer-wise
directional cosine with SFT is only $0.315/0.046/-0.038/0.088$ on
HotpotQA/2Wiki/MuSiQue/QASC, compared with
$0.789/0.469/0.769/0.660$ for the shared answer. The paired differences are
positive with non-overlapping bootstrap confidence intervals on all four
datasets (Appendix Table~\ref{tab:length-matched-gradient-control}).
\textbf{Shorter actions make token gradients cancel less, but shared answer
content is required to align the resulting update with SFT.} Coherence is
therefore a cancellation diagnostic, not a predictor of final F1.

\begin{findingbox}{Finding 3: Sequence Content Determines Directional Alignment}
Length explains much of the scalar coherence contraction, but not alignment
with the desired answer update; that alignment requires shared sequence content.
\end{findingbox}

\begin{table*}[t]
\centering
\caption{Three matched interventions locate where privileged OPD loses useful
supervision. \textbf{(a)} Reward rescaling is evaluated on the complete
HotpotQA and 2Wiki development sets. \textbf{(b)} At equal candidate and update counts,
answer-aware selection beats random selection on three datasets; displayed
deltas use unrounded scores.
\textbf{(c)} Projection-only changes the scored sequence, while ASA-OPD adds
answer-aware token selection (F1 / EM, \%). Each panel is an independent training block;
comparisons are valid within, not across, panels. Bold marks the strongest
nonbaseline intervention in each retrieval column of panel (c).}
\label{tab:intervention-summary}
\label{tab:asa-main}
\label{tab:asa-fallback-ablation}
\label{tab:advantage-normalization-full}
\footnotesize
\begin{minipage}[t]{0.43\textwidth}
\centering
\textbf{\textit{(a) Reward rescaling on HotpotQA (F1 / EM, \%)}}\par\vspace{3pt}
\setlength{\tabcolsep}{3.0pt}
\resizebox{\linewidth}{!}{%
\begin{tabular}{lccc}
\toprule
\textbf{Transform} & \textbf{Answer} & \textbf{Full evid.} & \textbf{Visible evid.} \\
\midrule
Raw & 49.91 / 31.09 & 33.28 / 8.01 & 37.32 / 12.99 \\
Centered & 48.31 / 29.05 & 34.66 / 9.57 & 36.87 / 12.44 \\
Standardized & \textbf{53.99 / 36.27} & \textbf{38.93 / 15.12} & \textbf{39.42 / 15.76} \\
\bottomrule
\end{tabular}
}

\vspace{5pt}
\textbf{\textit{2Wiki (BM25): full-evidence OPD (F1 / EM, \%)}}\par\vspace{2pt}
\begin{tabular*}{0.78\linewidth}{@{\extracolsep{\fill}}lc}
\toprule
\textbf{Transform} & \textbf{F1 / EM} \\
\midrule
Raw & 46.49 / 40.93 \\
Centered & 47.87 / 41.94 \\
Standardized & \textbf{48.00 / 41.98} \\
\bottomrule
\end{tabular*}
\end{minipage}
\hfill
\begin{minipage}[t]{0.54\textwidth}
\centering
\textbf{\textit{(b) Rate-matched token selection (F1, \%)}}\par\vspace{1pt}
\includegraphics[width=0.96\linewidth]{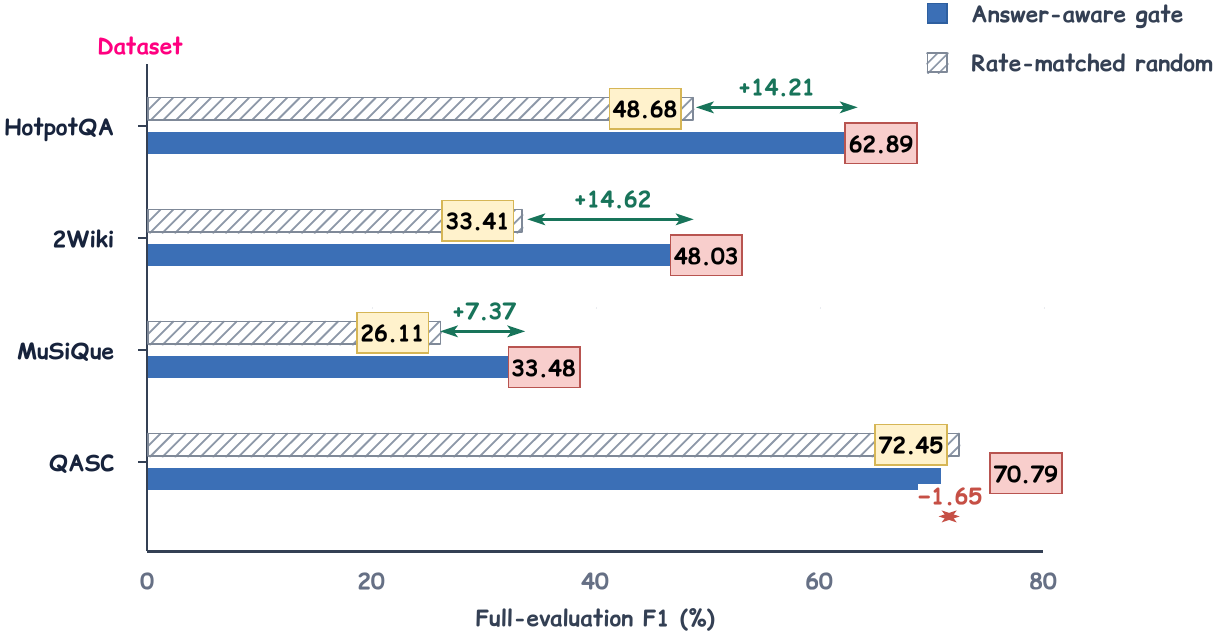}
\end{minipage}

\vspace{2pt}
\textbf{\textit{(c) Answer-space construction and anchoring}}\par\vspace{3pt}
\setlength{\tabcolsep}{1.2pt}
\begin{tabular*}{\textwidth}{@{\extracolsep{\fill}}lcccccccc}
\toprule
\multirow{2}{*}{\textbf{Training signal}}
& \multicolumn{2}{c}{\textbf{HotpotQA}} & \multicolumn{2}{c}{\textbf{2Wiki}}
& \multicolumn{2}{c}{\textbf{MuSiQue}} & \multicolumn{2}{c}{\textbf{QASC}} \\
\cmidrule(lr){2-3}\cmidrule(lr){4-5}\cmidrule(lr){6-7}\cmidrule(lr){8-9}
& \textbf{Qwen} & \textbf{BM25} & \textbf{Qwen} & \textbf{BM25} & \textbf{Qwen} & \textbf{BM25} & \textbf{Qwen} & \textbf{BM25} \\
\midrule
Full-evidence OPD & 66.02/52.61 & 57.86/45.43 & 52.88/45.90 & 46.49/40.93 & 28.88/17.17 & 14.09/6.66 & 20.90/18.03 & 8.80/6.91 \\
Projection only & 53.74/38.81 & 45.41/32.45 & 39.12/29.04 & 35.06/27.00 & 28.71/17.54 & 14.20/6.87 & 66.58/63.07 & 61.85/57.56 \\
Reference answer only & 42.66/27.49 & 36.60/23.43 & 42.65/34.41 & 43.96/37.35 & 31.46/19.74 & 19.28/10.38 & 62.47/59.07 & 69.45/67.82 \\
Evidence+answer weight & 56.40/41.20 & 39.13/26.31 & 36.39/27.47 & 31.26/23.42 & 30.06/18.33 & 15.23/7.45 & \textbf{71.21/69.33} & 69.25/67.71 \\
Answer-only weight & 56.64/41.51 & \textbf{50.21/37.50} & 44.81/36.35 & 40.97/34.41 & \textbf{37.68/25.73} & \textbf{19.89/11.25} & 70.75/68.90 & 69.65/68.25 \\
ASA-OPD & \textbf{62.89/48.67} & 48.22/35.40 & \textbf{48.03/39.93} & \textbf{44.08/37.72} & 33.48/21.89 & 19.51/10.80 & 70.79/68.57 & \textbf{69.88/68.47} \\
\bottomrule
\end{tabular*}
\end{table*}

\section{Interventions Locate Where Privileged Supervision Fails}
\label{sec:asa}

The analysis yields three matched interventions, each modifying a different
part of the update. Reward rescaling keeps the sequence and selected tokens
unchanged, testing whether calibration alone explains the teacher gap.
Projection-only replaces a valid free-form rollout with its extracted short
answer, testing whether the scored sequence is the bottleneck. ASA-OPD then
adds answer-aware token selection; a keep-rate-matched random gate tests whether
any gain comes from which tokens are retained rather than fewer updates.

\subsection{Reward Rescaling Does Not Resolve Unreliable Supervision}
On a matched HotpotQA control block, standardizing token weights improves
full-evidence OPD from $33.28\%$ to $38.93\%$ F1, but answer-only OPD also
improves from $49.91\%$ to $53.99\%$. The $15.06$-point teacher gap remains.
On the complete 2Wiki BM25 development set, centering and standardization improve
full-evidence OPD from $46.49\%$ to $47.87\%$ and $48.00\%$ F1, respectively.
\textbf{Reward scale affects optimization, but its modest and dataset-dependent
gains do not explain the teacher ordering}
(Table~\ref{tab:intervention-summary}a).

\subsection{Answer Projection Helps Only When Output Mismatch Dominates}

We separate sequence construction from token compatibility with two nested
interventions. \emph{Projection-only} trains a valid short answer extracted
from the student rollout, without reference-answer fallback or the
answer-only-teacher gate. \emph{ASA-OPD} constructs an answer-space candidate
and then retains only tokens supported by both teachers. Let $g(\hat y)$
extract a short answer from a student rollout. ASA's training candidate is
\[
u=
\begin{cases}
\operatorname{norm}(g(\hat y)), & g(\hat y)\text{ is valid},\\
\operatorname{norm}(a), & \text{otherwise},
\end{cases}
\]
where normalization lowercases text, removes punctuation and articles, and collapses whitespace. Extraction does not compare the candidate with $a$. The fallback path supplies a training-time candidate only when extraction fails; no gold answer or extractor is used at evaluation.

The rollout that produces $u$ is sampled on-policy, but projection changes the
scored sequence and its token prefixes. ASA is therefore an
\emph{on-policy-derived projected distillation} objective, rather than an
unbiased estimator of the original sampled-token objective. When extraction
fails, the reference fallback is explicitly a supervised training path.

Let $h_t^u=(x,u_{<t})$ denote the shared retrieved context and answer prefix. The evidence-and-answer and answer-only teachers score the same answer-space token $u_t$:
\[
\resizebox{0.96\columnwidth}{!}{$\displaystyle
A^{u}_{\mathrm{evid}}(t)
=\log\pi_T(u_t\mid h_t^u,z_{\mathrm{evid}})
-\log\pi_S(u_t\mid h_t^u).
$}
\]
\[
\resizebox{0.96\columnwidth}{!}{$\displaystyle
A^{u}_{\mathrm{ans}}(t)
=\log\pi_T(u_t\mid h_t^u,z_{\mathrm{ans}})
-\log\pi_S(u_t\mid h_t^u).
$}
\]
The answer-only teacher acts as an anchor rather than replacing the evidence-and-answer teacher. ASA retains an evidence-conditioned reward only when both teachers support the same answer token:
\[
\resizebox{0.82\columnwidth}{!}{$\displaystyle
m_t=\mathbf 1\!\left\{A^{u}_{\mathrm{evid}}(t)>0\wedge A^{u}_{\mathrm{ans}}(t)>0\right\}.
$}
\]
\[
\resizebox{0.96\columnwidth}{!}{$\displaystyle
\mathcal L_{\mathrm{ASA}}
=-\sum_t
\mathrm{sg}\!\left(m_tA^{u}_{\mathrm{evid}}(t)\right)
\log\pi_S(u_t\mid x,u_{<t}).
$}
\]
Both teachers rescore student-derived candidates under the projected prefix.
Controls isolate projection, reference-answer coverage, each teacher, and a
keep-rate-matched random gate. ASA uses the normalized reference answer only
when extraction fails during training; evaluation uses neither the extractor
nor the gold answer (Appendix~\ref{app:sensitivity}).

\paragraph{Increasing rollout coverage.}
Sections~\ref{sec:reachability}--\ref{sec:projection} show that longer
rollouts expose more correct and extractable answers. Extending training
rollouts from 32 to 128 tokens raises valid extraction from $71.4\%$ to
$82.6\%$, but changes Extract-only F1 only from $58.8\%$ to $59.3\%$ and
reduces ASA-OPD from $65.8\%$ to $63.1\%$ (Appendix
Figure~\ref{fig:extractor-full-eval}c).
Coverage is therefore necessary but not sufficient.

\paragraph{Effect of answer-space projection.}
The effect of projection is non-monotonic across datasets. Projection improves QASC by $45.68$
F1 points but reduces HotpotQA and 2Wiki by $12.28$ and $13.76$ points
(Table~\ref{tab:intervention-summary}c). Projection is beneficial when the free-form
rollout maps poorly to a short answer, but it can discard useful
sequence-level supervision when the original rollout already trains the task effectively.

\paragraph{Effect of answer-only-teacher anchoring.}
Adding answer-only-teacher support offsets part of the projection loss on HotpotQA
and 2Wiki and raises MuSiQue and QASC above projection alone. Across the two
retrieval states, ASA improves over projection alone in all eight comparisons,
but it remains below free-form full-evidence OPD on HotpotQA and 2Wiki. We
therefore use ASA as a mechanism intervention, not as a universally superior
training method.

\subsection{Answer-Conditioned Selection Matters Beyond Update Count}
We next keep the projected candidates and number of nonzero token updates
constant, changing only which tokens are retained. ASA exceeds the rate-matched
random gate by $14.21$, $14.62$, and $7.37$ F1 points on HotpotQA, 2Wiki, and
MuSiQue, respectively, but trails it by $1.65$ points on QASC
(Table~\ref{tab:intervention-summary}b). Thus, answer-aware token selection adds
value on three datasets; QASC is dominated by the projection step, where an
additional semantic gate removes beneficial token updates from the already
compact answer.

\paragraph{Intervention conclusion.}
Projection and compatibility-aware selection address different bottlenecks;
neither is uniformly beneficial across output spaces. Privileged information
helps only when it reaches the \textbf{task-relevant answer and token updates}.

\section{Related Work}

\paragraph{Retrieval-augmented and multi-hop QA.}
RAG conditions generation on retrieved text
\citep{guu2020retrieval,lewis2020retrieval,izacard2021leveraging,
borgeaud2022improving,izacard2023atlas,shi2024replug}; dynamic parametric RAG
converts documents into test-time parametric knowledge \citep{tan2025dynamic}.
Retrieval pipelines improve evidence through dense and end-to-end methods
\citep{karpukhin2020dense,singh2021end,izacard2021unsupervised}, multi-hop and
interleaved search \citep{khattab2021baleen,trivedi2023interleaving},
retrieval-aware generation \citep{zhang2024raft,asai2024self}, and reranking
\citep{nogueira2019passage,khattab2020colbert,nogueira2020document,
yu2024rankrag}. We ask whether teacher-only gold evidence improves
OPD supervision under a matched student retrieval state.

\paragraph{Privileged and on-policy distillation.}
Distillation transfers training-only knowledge at sequence, rationale, or preference level
\citep{vapnik2009new,hinton2015distilling,lopez2015unifying,
kim2016sequence,hsieh2023distilling,rafailov2023direct}. OPD instead scores
learner-induced states, often through reverse or skew divergences
\citep{ross2011reduction,lin2020autoregressive,agarwal2024policy,
gu2024minillm,ko2024distillm}. Recent work studies stabilization, token selection, and privileged traces
\citep{zhang2026full,pan2026rlcsd,li2026rethinking,zhu2026many,
wang2026demystifying,luo2026demystifying,plyusov2026trust,li2026filter,lazaridis2026edge,
zhang2026pi,yu2026dopd}; privileged self- and context distillation alter the supervision source
\citep{zhao2026self,penaloza2026privileged,ye2026policy,
yu2026dopd,kaur2026rethinking}.
We test when RAG evidence converts teacher judgment into useful
token-level supervision on student responses.

\section{Conclusion}

This work shows that privileged teacher information does not reliably improve
OPD in multi-hop RAG because reachability, projection, and compatibility can
prevent better judgment from becoming useful answer-token updates.
Matched-sequence analysis explains the OPD--SFT gap, while targeted
interventions reveal no universal correction, establishing a practical
principle: privilege must become answer-compatible supervision.

\section*{Limitations}

This study opens several directions for future work. The reachability--
projection--compatibility framework can be extended to long-form generation,
open-corpus retrieval, and multilingual settings. Learned answer projections
and adaptive compatibility estimators may further generalize the interventions,
while broader model and retrieval families can clarify how these supervision
requirements scale.

\section*{Ethical Considerations}

The experiments use publicly available question-answering benchmarks and
released pretrained models; no new human-subject or private personal data are
collected. Released examples may mention public entities from the source
benchmarks, but author and infrastructure identifiers are removed. Research and
downstream use should follow the applicable dataset and model licenses,
document computational resources, and evaluate factuality and fairness for the
intended setting. Applications in consequential domains should additionally
include domain-appropriate evaluation and human oversight.

\bibliography{references}

\clearpage
\phantomsection
\pdfbookmark[1]{Appendix}{appendix}
\appendix
\setlength{\textfloatsep}{7pt plus 1pt minus 2pt}
\setlength{\floatsep}{6pt plus 1pt minus 2pt}
\setlength{\intextsep}{6pt plus 1pt minus 2pt}
\setlength{\abovecaptionskip}{3pt}
\setlength{\belowcaptionskip}{2pt}
\renewcommand{\arraystretch}{0.94}
\raggedbottom
\renewcommand{\topfraction}{0.95}
\renewcommand{\bottomfraction}{0.85}
\renewcommand{\textfraction}{0.05}
\renewcommand{\floatpagefraction}{0.72}
\renewcommand{\dbltopfraction}{0.95}
\renewcommand{\dblfloatpagefraction}{0.72}
\makeatletter
\setlength{\@fptop}{0pt}
\setlength{\@fpsep}{7pt plus 2pt minus 2pt}
\setlength{\@fpbot}{0pt plus 1fil}
\setlength{\@dblfptop}{0pt}
\setlength{\@dblfpsep}{7pt plus 2pt minus 2pt}
\setlength{\@dblfpbot}{0pt plus 1fil}
\makeatother

\phantomsection
\pdfbookmark[2]{Appendix Roadmap}{appendix-roadmap}
\section*{Appendix Roadmap}
\begingroup
\scriptsize
\setlength{\fboxsep}{5pt}
\noindent\fcolorbox{tblHead}{tblBlue!18}{%
\begin{minipage}{0.94\linewidth}
\textbf{\textcolor{tblHead}{A. Reproducibility.}} Datasets, prompts,
optimization, extraction, answer-generator training, and reranker protocols.

\smallskip
\textbf{\textcolor{tblHead}{B. Phenomenon controls.}} Stronger teachers,
independent runs, cross-family full-development evaluation, prompt controls,
and generic token filters.

\smallskip
\textbf{\textcolor{tblHead}{C. Mechanism evidence.}} PLS/GA uncertainty,
reward calibration, outcome stratification, shared-answer and length-matched
gradients, all-trace one-step effects, and representation shifts.

\smallskip
\textbf{\textcolor{tblHead}{D. Boundary check.}} Evidence-supervised
reranking as a direct use of gold support labels.

\smallskip
\textbf{\textcolor{tblHead}{E. Intervention controls.}} Extraction, complete
ASA variants, fallback, objective, and budget ablations.
\end{minipage}}
\par\smallskip
\endgroup

\section{Reproducibility and Experimental Protocol}
\label{app:impl}

\subsection{Datasets}

\begin{itemize}
\setlength{\itemsep}{3pt}
\setlength{\parskip}{0pt}
\setlength{\parsep}{0pt}

\item \textbf{HotpotQA} is an open-answer benchmark built from Wikipedia
articles, with questions requiring evidence from multiple documents and
sentence-level supporting-fact annotations
\citep{yang2018hotpotqa}. The answer generator receives retrieved documents
containing candidate evidence and distractors. HotpotQA provides the primary
matched comparison of teacher privilege and the three-seed full-development
result.

\item \textbf{2WikiMultiHopQA} is an open-answer Wikipedia benchmark designed
to evaluate multi-hop reasoning together with supporting evidence and
intermediate reasoning paths \citep{ho2020constructing}. Its construction
combines structured Wikidata relations with unstructured Wikipedia text. We
use it to test whether the observed supervision behavior transfers beyond
HotpotQA under a separate question-construction process.

\item \textbf{MuSiQue} is an open-answer benchmark containing connected
two-to-four-hop questions constructed from composable single-hop questions
\citep{trivedi2022musique}. Its construction filters out examples whose hops
can be answered independently and includes distractor paragraphs. We use the
answerable split to test the analysis under longer evidence chains and a more
constrained notion of connected multi-hop reasoning.

\item \textbf{QASC} is an eight-choice grade-school science benchmark in which
the answer is obtained by composing two annotated facts retrieved from a large
sentence corpus \citep{khot2020qasc}. Unlike the other three datasets, every
answer candidate is already present in the question. We cast QASC as
generation by asking the model to output the selected option text, and use it
as a boundary case where privileged evidence disambiguates an existing closed
candidate set rather than introducing an unseen open-form answer.

\end{itemize}

\subsection{Prompt Construction}

For OPD, the student and teacher share the question, retrieved documents, and output instruction. Teacher prompts add one privileged field immediately before the question. The answer-only field contains the reference answer; the full-evidence field contains every annotated support plus the answer; and the visible-evidence field contains only supports whose document is already in the student's top-$k$ context. The templates below show the exact structure; braces denote example-specific fields.

\begin{center}
\setlength{\fboxsep}{5pt}
\fcolorbox{tblHead}{tblBlue!12}{%
\begin{minipage}{0.92\columnwidth}
\scriptsize\ttfamily\raggedright
Use the retrieved documents to answer the question.\\
Output only the answer. Do not explain.\\[2pt]
Document 1 (\{title\_1\}): \{text\_1\}\\
\ldots\\
Document $k$ (\{title\_$k$\}): \{text\_$k$\}\\[2pt]
\{teacher-only privileged field\}\\[2pt]
Question: \{question\}\\
Answer:
\end{minipage}}
\end{center}

\begin{center}
\setlength{\fboxsep}{5pt}
\fcolorbox{tblHead}{tblGreen!14}{%
\begin{minipage}{0.92\columnwidth}
\scriptsize\ttfamily\raggedright
Privileged Information (teacher only; use silently):\\
\textbf{Answer only:} Reference Answer: \{answer\}\\[2pt]
\textbf{Full evidence:} Evidence 1 (\{title\}): \{support\}\\
\ldots\\
Reference Answer: \{answer\}\\[2pt]
\textbf{Visible evidence:} the same field restricted to supports whose titles occur in the student context.
\end{minipage}}
\end{center}

\subsection{OPD Optimization}

We use on-policy student rollouts, score the sampled student tokens under the teacher, and train with the sampled-token objective in Section~\ref{sec:setup}. The log-ratio term is detached from the student update, so it acts as a sampled advantage on the student log-probability. We normalize by non-padding rollout tokens and ignore prompt tokens, padding tokens, and truncated continuation positions outside the sampled answer rollout. Main OPD runs score at most 32 newly generated rollout tokens per example. For rollout-length sensitivity, we additionally test 16, 64, and 96 generated tokens.

\subsection{ASA-OPD Extraction and Gating}

ASA-OPD first decodes the student rollout and applies a deterministic extractor that does not inspect the gold answer. It keeps text inside an explicit \texttt{<answer>} span when present; otherwise it takes the first non-empty answer line, strips prefixes such as ``Answer:'' or ``the answer is'', and removes terminal punctuation. A candidate is invalid if it is empty, exceeds the answer-space token limit, or contains explanation markers such as ``according to'', ``document'', ``because'', or ``based on''. Invalid candidates are skipped in the extract-only ablation. In default ASA-OPD training, they are replaced by the normalized reference answer so that the scored sequence remains in the answer space. This fallback is used only during training; evaluation uses the model's generated answer directly and never receives the reference answer.

\paragraph{Extract-only and dual fallback.}
Let $\tilde a=g(\hat y)$ be the short-answer candidate extracted from a student rollout. The \emph{extract-only} variant uses
\[
u_{\mathrm{ext}}=\operatorname{norm}(\tilde a)
\]
only when $\tilde a$ is valid; otherwise, the example is skipped and contributes no OPD update. This variant therefore tests answer-space projection using only content produced by the student.

The \emph{dual-fallback} variant, used by ASA-OPD, instead constructs
\[
u_{\mathrm{dual}}=
\begin{cases}
\operatorname{norm}(\tilde a), & \text{if }\tilde a\text{ is valid},\\
\operatorname{norm}(a), & \text{otherwise},
\end{cases}
\]
where $a$ is the reference answer. ``Fallback'' refers only to this training-time candidate substitution. ``Dual'' refers to the subsequent token gate: an evidence-and-answer-teacher advantage is retained only when the answer-only teacher also assigns positive advantage to the same token. Thus, dual fallback combines complete answer-space coverage with agreement between the two teachers; it does not expose the reference answer during evaluation.

\paragraph{Matched fallback-only control.}
This control performs the same online rollout and extraction test as ASA-OPD,
but applies unit-weight answer negative log-likelihood only to examples that
use the fallback path; extracted candidates receive zero loss. It therefore
matches ASA's selective reference-answer fallback while removing both teacher
advantages. All reported scores use the full evaluation set; Appendix
Table~\ref{tab:fallback-attribution} reports the matched attribution results.

The answer-space sequence is limited to 12 tokens. Prompt tokens, padding, and truncated continuation positions are ignored. For each answer-space token, the evidence-and-answer and answer-only teachers are scored on the same candidate sequence. The hard ASA gate keeps an evidence-conditioned reward only when both teachers assign positive sampled advantage to that token.
\emph{Keep} denotes a token that receives an ASA update. \emph{Drop} sets its ASA gradient to zero, either because the student already matches the teachers (\emph{saturation}) or because at least one teacher assigns it lower probability (\emph{conflict}).

\begin{figure*}[t]
\centering
\includegraphics[width=\textwidth]{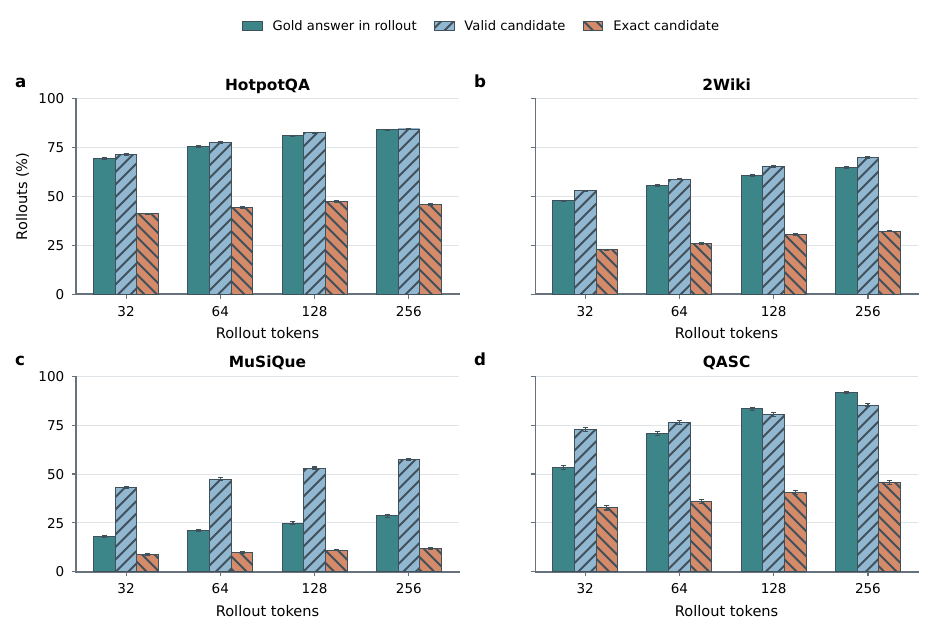}
\caption{Rollout-length analysis on the four full evaluation sets. Each
question has eight stochastic rollouts. Longer rollouts increase gold-answer
occurrence and valid extraction, while exact extraction improves more slowly
and can saturate. Error bars show bootstrap 95\% confidence intervals.}
\label{fig:rollout-length-full-eval}
\end{figure*}

Figure~\ref{fig:extractor-full-eval} reports full-evaluation training results
with the answer extractor. All four datasets use 128-token training rollouts;
HotpotQA additionally tests 32 tokens. Figure
\ref{fig:rollout-length-full-eval} separately characterizes how
rollout length changes answer occurrence and extraction across datasets.

\subsection{Answer Generator Training}

Unless otherwise specified, OPD, SFT, and reranker--answer-generator comparisons use a 5k-example training pool and 200 update steps. With four-way data parallel training, per-device batch size 1, and gradient accumulation 1, a 200-step run consumes 800 sampled training instances from the shuffled pool. The pool size controls sample diversity rather than the number of optimization steps.

The answer generator is Qwen3-14B with LoRA adaptation. LoRA uses rank 8 and alpha 16, targeting the attention and MLP projection matrices. The learning rate is $5\cdot10^{-6}$ and the maximum prompt length is 1024 tokens. External-teacher OPD uses Qwen3-32B only for scoring; the student remains Qwen3-14B.
Training uses NVIDIA A800 80GB and RTX 3090 24GB GPUs; all compared conditions
within an experimental block use the same hardware class and parallel
configuration. From completed runs with preserved runtimes and process counts,
we recover a conservative lower bound of 202.76 A800 and 872.32 RTX 3090
training GPU-hours (1,075.08 total). This excludes evaluation, interrupted runs,
and runs without preserved runtimes.

\paragraph{ASA scope and computation.}
ASA is designed as a mechanism intervention for short-answer RAG rather than
evidence-policy transfer for arbitrary long-form generation. Its two teacher
passes score at most 12 answer tokens. In matched 200-step HotpotQA runs, ASA
takes $20.4$ minutes versus $25.7$ minutes for full-evidence OPD. Broader output
spaces require learned or structured projections.

\subsection{Diagnostic and Sensitivity Runs}

The $D_{\text{all}}$ diagnostic is document-level: it inserts gold-document containers using the same truncation policy as other student contexts. We separately measure support-sentence survival and use a sentence-packed control with exact support preservation. For prompt-control runs, we evaluate the frozen teacher on the diagnostic set, reporting teacher answer F1, EM, output length, clean-answer rate, and answer log-probability before training the student. We include sensitivity checks over rollout length, sampling temperature/top-$p$, and sampled-advantage clipping.

The token-conflict analysis uses predefined lexical categories. Answer tokens occur in the normalized gold answer; evidence tokens occur in annotated support text; and discourse markers belong to \{\texttt{according}, \texttt{because}, \texttt{therefore}, \texttt{evidence}, \texttt{document}, \texttt{fact}, \texttt{based}, \texttt{given}, \texttt{so}, \texttt{thus}, \texttt{answer}, \texttt{question}\}. These categories record token provenance and explicit discourse cues; they are not model-predicted reasoning labels.

Layer-wise coherence is computed separately for each example and objective.
After one backward pass, $g_{\ell,t}$ is the gradient with respect to the
layer-$\ell$ hidden state at token position $t$; token gradients are combined
only within that example, and confidence intervals resample examples. It is
therefore independent of LoRA-module scaling. The objective-specific comparison
uses the sequence optimized by each objective: the normalized gold answer for
direct supervision, the projected answer candidate for ASA-OPD, and the
free-form student rollout for full-evidence OPD. The shared-answer control
instead scores the same normalized gold-answer token sequence under all three
objectives. This fixes the answer prefix, token identities, and token count,
so the remaining differences reflect the objective and teacher conditioning
rather than different sequences.

For retrieval context-size sensitivity, we retrain the answer generator on top-$5$ and top-$10$ reranked contexts. Top-$10$ uses a longer maximum sequence length but shorter per-document truncation so the comparison remains within the answer generator context window.

\subsection{Reranker Training and Evaluation}

The reranking comparison uses Qwen3-Reranker-8B in two conditions with the same base weights. The off-the-shelf condition uses the released model without task-specific adaptation. The evidence-supervised condition loads a LoRA adapter trained on 60,000 pointwise question--document pairs constructed from 5,000 training questions from each of HotpotQA, 2Wiki, and MuSiQue. Gold supporting documents receive the label \texttt{yes}; non-supporting candidates from the same question receive \texttt{no}. The model is trained to classify the final \texttt{yes}/\texttt{no} decision token with cross-entropy, and inference ranks documents by the normalized probability of \texttt{yes}.

Reranker training uses 300 optimizer steps, learning rate $10^{-5}$, per-device batch size 2, gradient accumulation 8, and maximum sequence length 768. LoRA uses rank 16, alpha 32, dropout 0.05, and targets the query, key, value, and output attention projections. All pairs use the shared instruction: \emph{Given a multi-hop question, identify passages containing evidence needed to answer it.}

HotpotQA and 2Wiki provide ten candidate documents per question. MuSiQue provides a larger paragraph pool, from which we keep the retrieved candidate list used by the open-book setup. During pair construction, each question contributes all available positive supports and between two and six candidate negatives; document text is truncated to 700 characters.

We do not insert gold documents into evaluation candidate pools. The achievable hit-all@5 is therefore bounded by the candidate-pool oracle, and retrieval improvements should be read as better ordering within the available pool rather than recovery of documents absent from the pool.

The reranker is trained only on training examples; evaluation examples are disjoint. Each reranker selects five documents. For every retrieval condition, a separate Qwen3-14B answer generator receives direct answer supervision on contexts produced by that condition and is evaluated on its corresponding full evaluation set. Training and evaluation contexts therefore follow the same retrieval distribution within each condition. Answer evaluation uses the standard normalized token-level F1 and exact match used in HotpotQA-style QA.

\section{Additional Teacher and Prompt Controls}
\label{app:additional-opd}

\subsection{Stronger Teacher and Seed Variation}

Same-model self-distillation could exaggerate the failure if the model's own
evidence-conditioned distribution is unusually misaligned. We therefore use a
frozen Qwen3-32B teacher to score Qwen3-14B student rollouts. In the
representative HotpotQA run in Table~\ref{tab:teacher-privilege}, answer-only
OPD reaches 60.9\% F1 and full-evidence OPD reaches 58.7\%, while direct answer
supervision under the same retrieval state reaches 74.8\%. Sensitivity results
in Table~\ref{tab:opd-sensitivity} reverse the ordering in several
configurations, so this control supports the gap to direct supervision rather
than a universal ranking between the two teachers.

The prompt controls separate teacher accuracy from distillation usefulness. For example, the XML evidence-and-answer teacher reaches 93.1\% frozen-teacher F1, yet its student reaches only 40.9\% F1 after OPD. A teacher that answers correctly is not necessarily a good token-level teacher.

The same-model comparison in Table~\ref{tab:teacher-privilege} is
stable across three training seeds. This check supports the phenomenon
without changing its scope: it concerns same-model sampled-token OPD under a
common BM25 top-$5$ student context.

The filtering controls below show the same boundary from another angle. Token-level filtering improves evidence OPD, but direct answer supervision remains the strongest objective under the same retrieved context.

\subsection{Cross-Family Full-Development Evaluation}

We evaluate independently trained Llama-3.2-3B checkpoints on each complete
development set. Table~\ref{tab:cross-family-full-dev} reports paired bootstrap
intervals over evaluation examples. Because each checkpoint uses one training
seed, these intervals quantify evaluation-set uncertainty rather than training
variance.

\begin{table}[!htbp]
\centering
\caption{Cross-family evaluation on complete development sets (F1, \%).
$\Delta$ is full-evidence minus answer-only OPD with a paired 95\% bootstrap
CI. Bold marks the strongest objective.}
\label{tab:cross-family-full-dev}
\scriptsize
\setlength{\tabcolsep}{2.2pt}
\resizebox{\columnwidth}{!}{%
\begin{tabular}{lrrrrr}
\toprule
\textbf{Dataset} & \textbf{$n$} & \textbf{Answer} & \textbf{Full evid.} &
\textbf{$\Delta$ [95\% CI]} & \textbf{SFT} \\
\midrule
HotpotQA & 7,405  & 64.32 & 64.37 & +0.05 [\textminus0.17, 0.27] & \textbf{66.28} \\
2Wiki    & 12,576 & 34.61 & 35.24 & +0.63 [0.28, 0.97]           & \textbf{47.09} \\
MuSiQue  & 2,417  & 34.65 & 28.77 & \textminus5.88 [\textminus6.95, \textminus4.83] & \textbf{38.00} \\
\bottomrule
\end{tabular}}
\end{table}

Full evidence is indistinguishable from answer-only supervision on HotpotQA,
slightly better on 2Wiki, and substantially worse on MuSiQue. The exact
teacher ordering therefore changes across datasets, while SFT remains strongest
throughout. This second model family supports the paper's central conclusion:
teacher information quantity alone does not determine supervision utility.

\subsection{Prompt-Level and Filtering Controls}

Table~\ref{tab:prompt-controls} evaluates the teacher's own quality. The XML answer-format teacher achieves 93.1\% F1 on its own frozen output, yet student OPD trained from it reaches only 40.9\% F1. This makes poor teacher answer accuracy or raw output format insufficient explanations. No prompt-level control we tested makes evidence privilege a dependable signal for the answer generator.

\begin{table}[!htbp]
\centering
\caption{HotpotQA prompt controls on the diagnostic set. Teacher F1 is
measured before training; student F1/EM is measured after OPD. Accurate
evidence-conditioned teachers can still provide weak token-level supervision.}
\label{tab:prompt-controls}
\scriptsize
\setlength{\tabcolsep}{1.5pt}
\resizebox{\columnwidth}{!}{%
\begin{tabular}{p{0.45\columnwidth}ccc}
\toprule
\textbf{Prompt control} & \textbf{Teach. F1 (\%)} & \textbf{Stud. F1 (\%)} & \textbf{Stud. EM (\%)} \\
\midrule
Random distractors & 55.2 & 33.7 & 4.2 \\
Duplicated docs & 52.6 & 30.9 & 0.8 \\
Highlighted support & 51.4 & 50.3 & 29.0 \\
Ordered evidence & 60.3 & 49.7 & 27.4 \\
XML evidence answer & \textbf{93.1} & 40.9 & 15.2 \\
\bottomrule
\end{tabular}
}
\end{table}

The full filtering analysis includes teacher-top-$20$ JSD and a sample-level hit-all gate in addition to the token mask summarized in the main text. Neither distributional matching nor sample-level filtering recovers direct answer supervision under the same retrieval state.

\begin{figure}[t]
\centering
\includegraphics[width=\columnwidth]{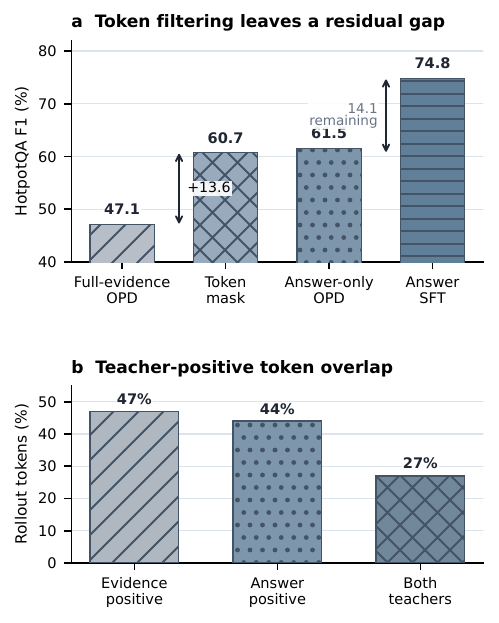}
\caption{Generic evidence-side filters on the predefined HotpotQA diagnostic
set. \textbf{(a)} Positive-advantage token masking recovers 13.6 F1 points but
retains a 14.1-point gap to direct answer supervision; double arrows mark both
differences. \textbf{(b)} Evidence-positive and answer-positive tokens overlap
on only $27\%$ of the rollout. ASA-OPD applies agreement filtering after
projecting the rollout into the answer space.}
\label{fig:opd-filtering-controls}
\end{figure}

\section{Mechanism Diagnostics}
\label{app:diagnostics}

\begin{figure*}[t]
\centering
\includegraphics[width=0.92\textwidth]{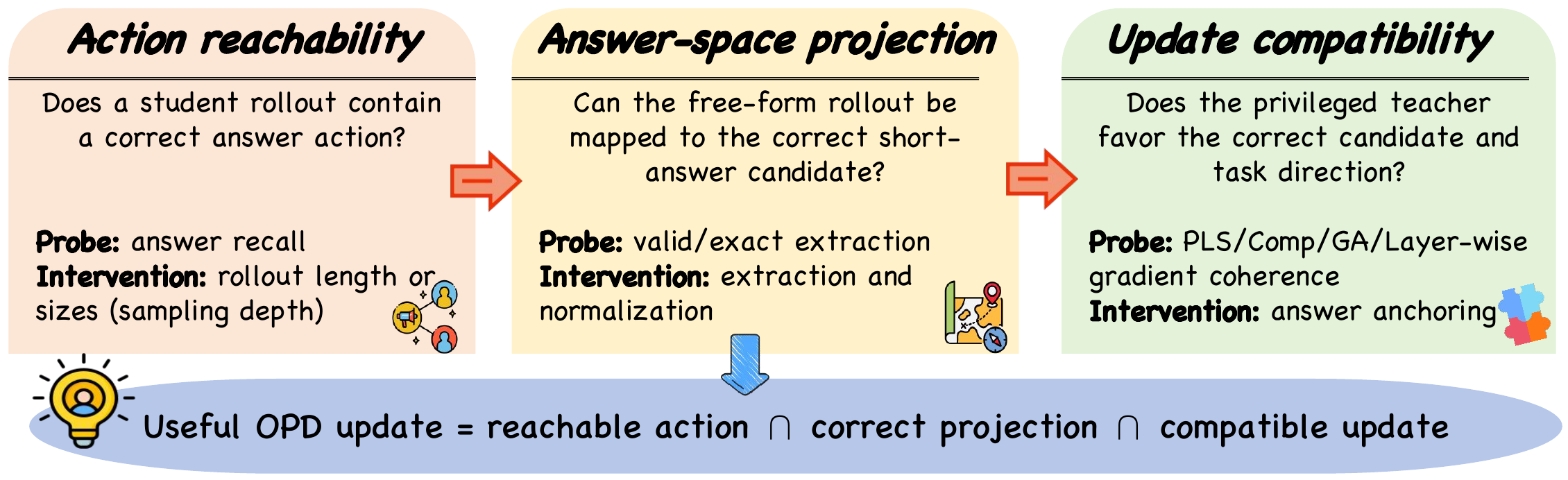}
\caption{Three requirements for useful privileged OPD supervision. The student
rollout must reach a usable answer, projection must recover it, and the teacher
weights must reinforce its tokens; direct answer supervision supplies the
target sequence and positive weights directly.}
\label{fig:three-stage-detail}
\end{figure*}

Figure~\ref{fig:three-stage-detail} expands the three requirements used to
organize the mechanism analysis.

\subsection{Diagnostic Definitions}

Privilege-induced log-probability shift measures how privilege $z$ changes
teacher support on a student rollout:
\[
\resizebox{0.96\columnwidth}{!}{$\displaystyle
\mathrm{PLS}(z)=\frac{1}{T}\sum_t
\left[\log\pi_T(\hat y_t\mid x,z,\hat y_{<t})
-\log\pi_T(\hat y_t\mid x,\hat y_{<t})\right].
$}
\]
Let \(\bar A_z(\hat y)=T^{-1}\sum_t A_t(z)\) and let
\(\mathrm{Ans}(\hat y)\) be the normalized extracted answer. Outcome
separation is
\[
\resizebox{0.96\columnwidth}{!}{$\displaystyle
\mathrm{Comp}(z)=
\mathbb E[\bar A_z(\hat y)\mid \mathrm{Ans}(\hat y)=a]
-\mathbb E[\bar A_z(\hat y)\mid \mathrm{Ans}(\hat y)\neq a].
$}
\]
Finally, for the loss gradient $g_{\ell,t}$ at token position $t$ in layer
$\ell$, token-gradient coherence is
\[
C_\ell=
\frac{\left\|\sum_t g_{\ell,t}\right\|_2}
{\sum_t\left\|g_{\ell,t}\right\|_2}.
\]
Higher coherence indicates that token-level updates reinforce rather than
cancel one another.

\begin{figure*}[t]
\centering
\includegraphics[width=0.82\textwidth]{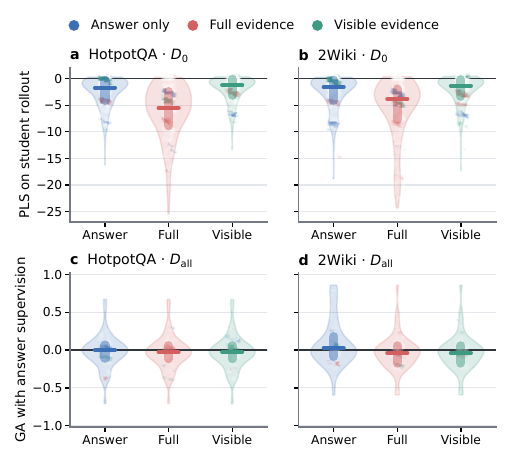}
\caption{Privilege-induced shifts and answer-gradient alignment on the
diagnostic sets. PLS measures how teacher privilege changes support for the
student rollout; GA measures whether the resulting update follows the direct
answer-supervision direction. Thick intervals show bootstrap 95\% confidence
intervals.}
\label{fig:pls-ga-mechanism}
\end{figure*}

\subsection{Stability and Layer-Wise Diagnostics}

Table~\ref{tab:diagnostic-ci} tests whether the PLS and GA patterns survive
example-level resampling. The intervals preserve the central distinction:
teacher support for a rollout and alignment with direct answer supervision are
different quantities.

\begin{table*}[t]
\centering
\caption{Bootstrap confidence intervals for PLS and GA on the diagnostic
sets. PLS is stratified by rollout correctness; GA compares matched questions
under the same student state.}
\label{tab:diagnostic-ci}
\scriptsize
\setlength{\tabcolsep}{3pt}
\begin{tabular*}{\textwidth}{@{\extracolsep{\fill}}llcccc}
\toprule
\textbf{Dataset} & \textbf{State} & \textbf{Answer PLS} & \textbf{Full-evidence PLS} & \textbf{Answer GA} & \textbf{Full-evidence GA} \\
\midrule
HotpotQA & $D_0$ & \textminus2.60 [\textminus2.98, \textminus2.21] & \textminus6.24 [\textminus6.95, \textminus5.52] & +0.045 [\textminus0.012, +0.104] & +0.017 [\textminus0.029, +0.062] \\
HotpotQA & $D_{\mathrm{all}}$ & \textminus1.47 [\textminus1.77, \textminus1.16] & \textminus1.54 [\textminus1.88, \textminus1.21] & \textminus0.020 [\textminus0.082, +0.043] & \textminus0.033 [\textminus0.095, +0.032] \\
2Wiki & $D_0$ & \textminus2.76 [\textminus3.22, \textminus2.33] & \textminus5.79 [\textminus6.57, \textminus5.02] & +0.013 [\textminus0.045, +0.075] & +0.007 [\textminus0.051, +0.068] \\
2Wiki & $D_{\mathrm{all}}$ & \textminus1.54 [\textminus1.86, \textminus1.26] & \textminus1.86 [\textminus2.22, \textminus1.53] & +0.087 [+0.005, +0.174] & \textminus0.031 [\textminus0.094, +0.037] \\
\bottomrule
\end{tabular*}
\end{table*}

\paragraph{Sequence-matched gradients.}

\begin{figure*}[t]
\centering
\includegraphics[width=\textwidth]{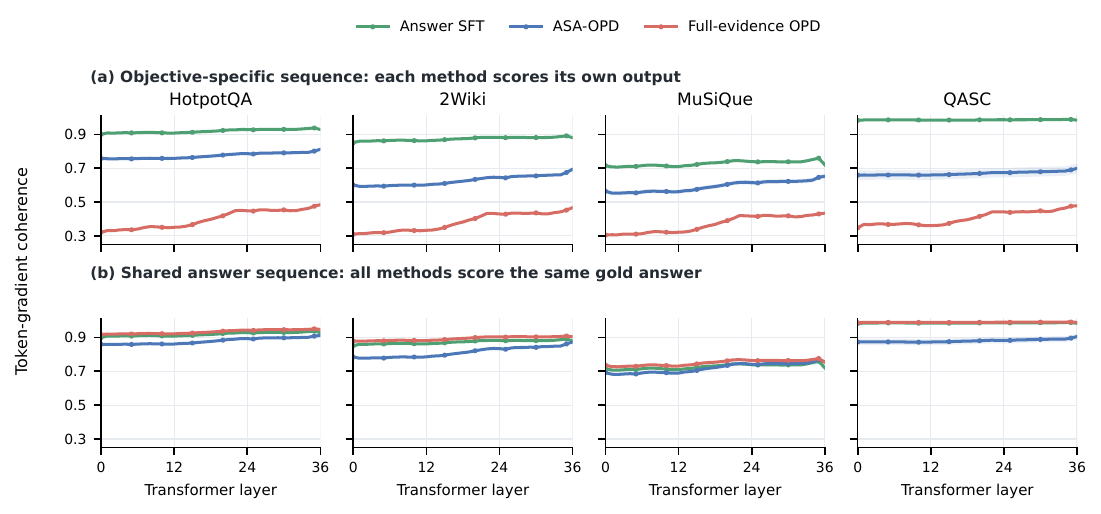}
\caption{Layer-wise token-gradient coherence on the four full evaluation sets.
The top row scores each objective's own training sequence; the bottom row
scores the same normalized gold-answer sequence under all objectives. Bands
show bootstrap 95\% confidence intervals.}
\label{fig:layerwise-gradient-coherence}
\end{figure*}

Figure~\ref{fig:layerwise-gradient-coherence} separates sequence identity from
objective identity. The large discrepancy appears when each objective scores
its own sequence and largely disappears when all objectives score the same
normalized answer.

Table~\ref{tab:length-matched-gradient-control} shows that the length-matched
spans substantially reduce the coherence gap, confirming that this scalar
statistic is length-sensitive. Yet their update directions remain far from
answer SFT. Shared-answer scoring improves the paired cosine on every dataset,
so equal token count cannot explain the directional result.

\paragraph{All-trace one-step effect.}
We apply equal-norm updates to all 472 teacher-selected HotpotQA traces whose
rollout contains the correct answer. Table~\ref{tab:one-step-all-traces}
reports the resulting change in answer NLL; negative values indicate an
immediate improvement. Both the full-sequence and positive-answer-token
updates lower answer NLL, but the latter produces more than twice the mean
reduction. Its paired advantage over the full sequence is $0.01244$ (95\% CI
$[0.00708,0.01783]$). Thus the corrected interpretation is dilution of useful
answer credit by the rest of the sampled sequence, not absence of any useful
full-sequence update.
This local intervention tests the immediate direction of an update and is not
treated as a predictor of final benchmark F1.

\paragraph{Matched-answer representations.}

\begin{figure*}[t]
\centering
\includegraphics[width=0.80\textwidth]{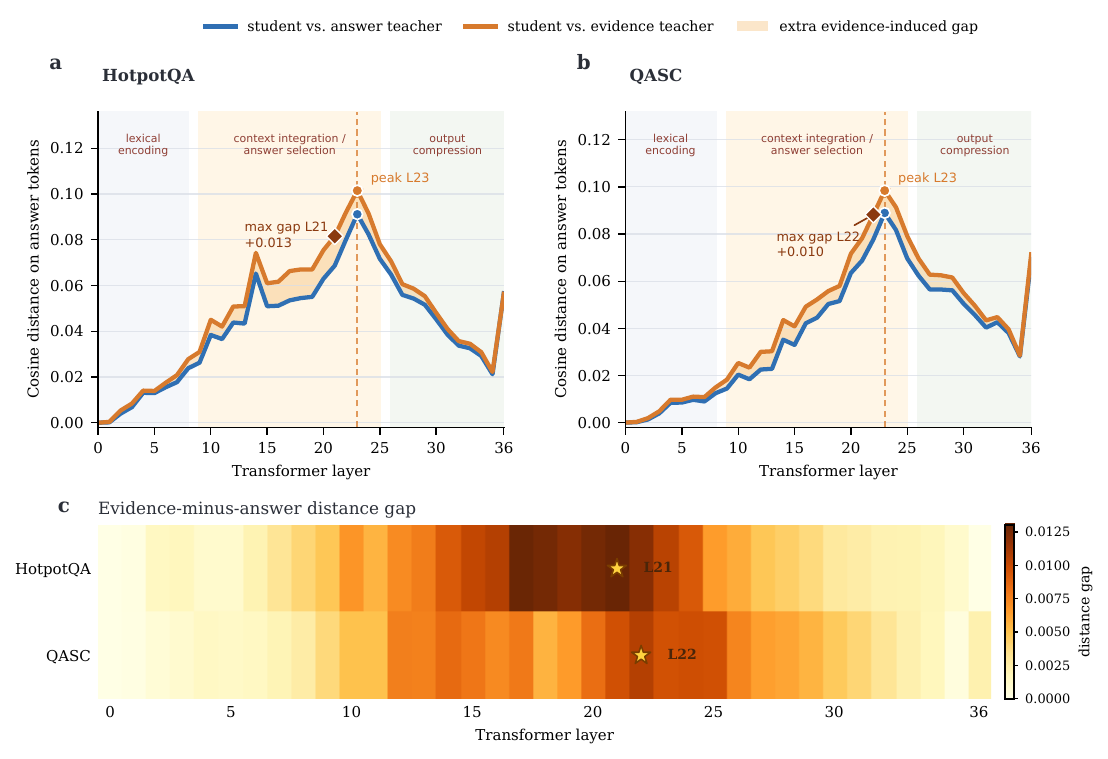}
\caption{Layer-wise answer-token representation divergence under a matched
answer sequence. Lower panels show the evidence-minus-answer gap.}
\label{fig:layer-divergence-details}
\end{figure*}

Figure~\ref{fig:layer-divergence-details} gives a descriptive layer-wise view
using the same answer-token sequence in every condition. Evidence-induced
separation is small in early layers and larger in middle layers on both
datasets. Because this probe is representational rather than interventional,
we use it only to localize the shift, not to infer a causal computation stage.

\clearpage
\twocolumn[{
\begin{minipage}{\textwidth}
\centering
\captionof{table}{Controls separating sequence length, content, and update direction.
\textbf{(a)} Equal-length rollout spans are compared with shared-answer
scoring; $n$ excludes one-token answers. \textbf{(b)} Equal-norm updates are
evaluated on all 472 correct teacher-selected HotpotQA traces. Negative
$\Delta$NLL indicates improvement.}
\label{tab:length-matched-gradient-control}
\label{tab:one-step-all-traces}
\small
\setlength{\tabcolsep}{7pt}
\renewcommand{\arraystretch}{1.12}
\textbf{\textit{(a) Length-matched gradient control}}
\par\vspace{4pt}
\begin{tabular*}{0.98\textwidth}{@{\extracolsep{\fill}}lrrrr}
\toprule
\textbf{Dataset} & \textbf{$n$} &
\textbf{Gap closed: LM / shared} & \textbf{Cosine: LM / shared} &
\textbf{Shared--LM $\Delta$ [95\% CI]} \\
\midrule
HotpotQA & 455 & 59.5--81.0 / 96.5 & 0.315 / \textbf{0.789} & 0.474 [0.421, 0.527] \\
2Wiki    & 430 & 55.7--72.0 / 94.6 & 0.046 / \textbf{0.469} & 0.423 [0.349, 0.491] \\
MuSiQue  & 501 & 55.4--57.0 / 93.1 & \textminus0.038 / \textbf{0.769} & 0.808 [0.755, 0.858] \\
QASC     & 419 & 66.4--82.4 / 99.0 & 0.088 / \textbf{0.660} & 0.572 [0.497, 0.643] \\
\bottomrule
\end{tabular*}

\vspace{9pt}
\textbf{\textit{(b) All-trace one-step answer-loss effect}}
\par\vspace{4pt}
\begin{tabular*}{0.88\textwidth}{@{\extracolsep{\fill}}lcc}
\toprule
\textbf{OPD update} & \textbf{$\Delta$ answer NLL [95\% CI]} &
\textbf{Cosine with SFT [95\% CI]} \\
\midrule
Full sequence & \textminus0.01081 [\textminus0.01712, \textminus0.00443]
& 0.061 [0.019, 0.105] \\
Signed answer span & \textminus0.00549 [\textminus0.01315, 0.00225]
& \textminus0.056 [\textminus0.129, 0.016] \\
Positive answer tokens & \textbf{\textminus0.02325 [\textminus0.02809, \textminus0.01876]}
& \textbf{0.504 [0.468, 0.541]} \\
\bottomrule
\end{tabular*}

\vspace{3pt}
\begin{minipage}{0.94\textwidth}
\footnotesize
LM reports the strongest of prefix, suffix, and seeded-random equal-length
windows; the coherence entry is their range. In panel (b), the
positive-answer-token cosine uses 404 traces with a nonzero retained gradient;
all NLL statistics use 472 traces.
\end{minipage}
\end{minipage}
\vspace{8pt}
}]

\section{Evidence-Supervised Reranking as a Boundary Check}
\label{sec:component-actionable}

The main analysis studies gold evidence as privileged context for an OPD
teacher. Here we test a direct use of the same supervision: gold supports label
the documents that a reranker should select. We train Qwen3-Reranker-8B on
support--distractor pairs, select five documents, and train the answer generator
on that retrieval state (Appendix~\ref{app:impl}). Evidence training improves
over the same off-the-shelf reranker on all three full evaluation sets
(Table~\ref{tab:routing-main}), with larger gains on 2Wiki and MuSiQue, where a
required support is more often initially excluded.

Answer gains concentrate where reranking recovers a missing support and are
small when recall is unchanged. Evidence must pass through sampling,
projection, and token weighting in answer generation, but directly identifies
the training target in reranking.

\section{Intervention and Objective Controls}
\label{app:sensitivity}

\subsection{Extraction, Token Support, and Objective Controls}

\begin{figure*}[t]
\centering
\begin{minipage}{\textwidth}
\centering
\includegraphics[width=0.92\linewidth]{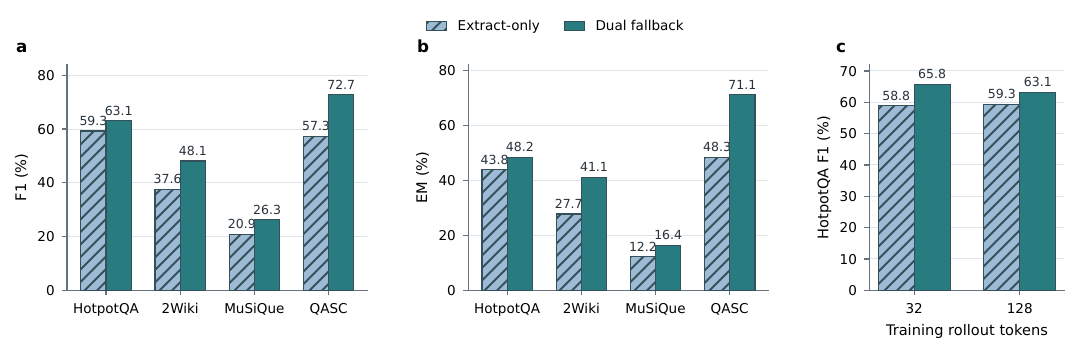}
\captionof{figure}{Full-evaluation performance with the answer extractor.
\textbf{(a--b)} F1 and EM use 128-token training rollouts; ASA-OPD improves
over Extract-only on all four datasets. \textbf{(c)} On HotpotQA, extending
training rollouts from 32 to 128 tokens does not improve either variant.}
\label{fig:extractor-full-eval}
\end{minipage}
\end{figure*}

The extractor maps a free-form rollout to a normalized short-answer
candidate. Across the four full evaluation sets, ASA-OPD improves over
Extract-only after this projection (Figure~\ref{fig:extractor-full-eval}).
Longer rollouts make the gold answer easier to reach and recover, but do not
improve the resulting HotpotQA model. Better answer coverage is therefore
useful but insufficient; the projected answer must also produce a compatible
update.

\paragraph{Optimization and objective controls.}
Table~\ref{tab:opd-sensitivity} varies rollout length and scoring
hyperparameters. The ordering among OPD variants changes across settings, so
the claim is not that answer-only OPD always dominates. The stable pattern is
that evidence-conditioned OPD is sensitive and remains below direct answer
supervision under the same retrieval state.

\begin{table*}[t]
\centering
\caption{Evidence-supervised reranking improves evidence coverage and
downstream QA. Retrieval metrics use matched diagnostic sets; F1 and EM use
the full evaluation sets.}
\label{tab:routing-main}
\small
\setlength{\tabcolsep}{7pt}
\begin{tabular}{llrrrr}
\toprule
\textbf{Dataset} & \textbf{Retrieval state} & \textbf{hit-all@5} & \textbf{recall@5} & \textbf{F1} & \textbf{EM} \\
\midrule
\multirow{3}{*}{HotpotQA}
& BM25 & 64.45 & 81.40 & 58.51 & 43.67 \\
& Qwen & 94.85 & 97.35 & 72.11 & 57.81 \\
& Evidence-trained & \textbf{98.20} & \textbf{99.05} & \textbf{72.63} & \textbf{58.22} \\
\midrule
\multirow{3}{*}{2Wiki}
& BM25 & 53.85 & 78.55 & 49.04 & 42.70 \\
& Qwen & 83.80 & 94.29 & 55.64 & 47.88 \\
& Evidence-trained & \textbf{97.15} & \textbf{98.80} & \textbf{58.05} & \textbf{50.33} \\
\midrule
\multirow{3}{*}{MuSiQue}
& BM25 & 16.65 & 50.34 & 23.15 & 13.53 \\
& Qwen & 61.50 & 83.91 & 43.53 & 30.99 \\
& Evidence-trained & \textbf{68.55} & \textbf{86.93} & \textbf{46.58} & \textbf{33.97} \\
\bottomrule
\end{tabular}
\end{table*}

\begin{table*}[t]
\centering
\caption{Extractor behavior and fallback attribution. \textbf{(a)} Exact is
the fraction of all rollouts whose extracted candidate matches the normalized
reference answer; precision is Exact among valid candidates. \textbf{(b)} All
F1 scores use the Qwen retrieval state and the same corrected answer parser as
Table~\ref{tab:intervention-summary}.}
\label{tab:extractor-accuracy}
\label{tab:fallback-attribution}
\small
\setlength{\tabcolsep}{7pt}
\textbf{\textit{(a) Extractor validity and correctness (\%)}}\par\vspace{3pt}
\begin{tabular}{lrrrrrr}
\toprule
& \multicolumn{3}{c}{\textbf{32-token rollout}} & \multicolumn{3}{c}{\textbf{128-token rollout}} \\
\cmidrule(lr){2-4}\cmidrule(lr){5-7}
\textbf{Dataset} & \textbf{Valid} & \textbf{Exact} & \textbf{Precision} & \textbf{Valid} & \textbf{Exact} & \textbf{Precision} \\
\midrule
HotpotQA & 71.4 & 41.1 & 57.5 & 82.6 & 47.3 & 57.2 \\
2Wiki    & 53.0 & 22.8 & 43.1 & 65.3 & 30.5 & 46.7 \\
MuSiQue  & 43.1 & 8.7  & 20.3 & 52.9 & 11.1 & 21.0 \\
QASC     & 72.6 & 32.7 & 45.0 & 80.5 & 40.5 & 50.3 \\
\bottomrule
\end{tabular}

\vspace{5pt}
\textbf{\textit{(b) Matched fallback attribution (F1, \%)}}\par\vspace{3pt}
\setlength{\tabcolsep}{8pt}
\begin{tabular}{lrrrr}
\toprule
\textbf{Dataset} & \textbf{Train fallback} & \textbf{Fallback-only} & \textbf{Random gate} & \textbf{ASA} \\
\midrule
HotpotQA & 28.5 & 40.56 & 48.68 & \textbf{62.89} \\
2Wiki    & 21.5 & 27.65 & 33.41 & \textbf{48.03} \\
MuSiQue  & 32.3 & 14.69 & 26.11 & \textbf{33.48} \\
QASC     & 25.8 & 65.36 & \textbf{72.45} & 70.79 \\
\bottomrule
\end{tabular}
\end{table*}

Validity therefore overestimates semantic correctness: at 128 tokens, only
$21.0$--$57.2\%$ of valid candidates are exact. Projection reduces the output
space but does not itself identify the correct answer, which explains why
coverage and extraction gains need not improve the trained model. Train
fallback is the mean fraction of examples using the normalized reference
answer over 200 updates. Fallback is used on $21.5$--$32.3\%$ of examples and
is therefore a substantial supervised path. ASA exceeds fallback-only on all
four datasets, so reference substitution alone does not explain its
performance. With fallback construction and update count held constant, ASA
also exceeds the random gate on the three open-answer datasets, but not on
QASC; the gating claim is restricted accordingly.

\begin{figure*}[t]
\centering
\begin{minipage}{0.96\textwidth}
\centering
\includegraphics[width=\linewidth]{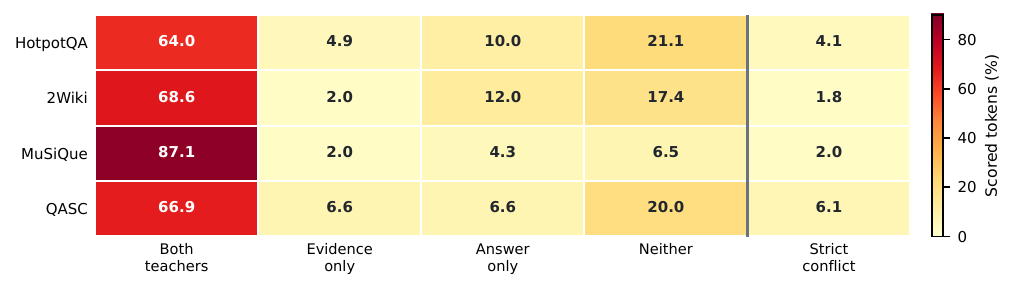}
\captionof{figure}{Answer-space teacher support on the diagnostic sets. The first four
columns partition tokens supported by both teachers, only the evidence
teacher, only the answer-only teacher, or neither. Strict conflict denotes positive
evidence advantage paired with negative answer advantage. ASA retains
evidence rewards supported by the answer-only teacher and removes
evidence-only updates, separating answer-space alignment from extraction.}
\label{fig:asa-token-diagnostic}
\end{minipage}
\end{figure*}

\begin{table*}[t]
\centering
\caption{Optimization and context controls. Panel (a) varies rollout and
scoring choices on the HotpotQA diagnostic set. Panel (b) verifies that
support survives prompt construction. Panel (c) compares gradient advantage
(GA) after preserving every annotated support sentence.}
\label{tab:optimization-controls}
\label{tab:opd-sensitivity}
\label{tab:support-visibility}
\label{tab:strict-matched-state}
\small
\setlength{\tabcolsep}{7pt}
\renewcommand{\arraystretch}{1.08}
\textbf{\textit{(a) Rollout and scoring sensitivity (F1 / EM, \%)}}\par\vspace{3pt}
\begin{tabular*}{0.90\textwidth}{@{\extracolsep{\fill}}lccc}
\toprule
\textbf{Setting} & \textbf{Answer only} & \textbf{Full evidence} & \textbf{Visible evidence} \\
\midrule
Rollout 16 & 60.8 / 43.8 & 59.3 / 41.8 & 60.6 / 43.0 \\
Rollout 64 & 56.1 / 38.8 & 58.1 / 40.6 & 58.3 / 41.2 \\
Rollout 96 & 55.1 / 37.8 & 55.6 / 37.4 & 57.6 / 40.8 \\
\midrule
Temperature 0.7 & 53.6 / 34.0 & 48.5 / 26.6 & 49.7 / 28.6 \\
Top-$p$ 0.80 & 47.2 / 23.4 & 45.8 / 21.6 & 45.5 / 21.4 \\
Advantage clip 0.5 & 44.4 / 20.2 & 46.6 / 22.8 & 45.6 / 21.4 \\
Advantage clip 1.0 & 45.0 / 20.6 & 46.4 / 23.2 & 45.4 / 20.6 \\
Advantage clip 2.0 & 47.1 / 24.2 & 45.3 / 22.2 & 45.3 / 20.8 \\
\bottomrule
\end{tabular*}

\vspace{8pt}
\textbf{\textit{(b) Support visibility after prompt construction}}\par\vspace{3pt}
\begin{tabular*}{0.68\textwidth}{@{\extracolsep{\fill}}lrrr}
\toprule
\textbf{Control} & \textbf{Full sentence} & \textbf{All visible} & \textbf{Token recall} \\
\midrule
2Wiki $D_{\text{all}}$ & 97.4\% & 93.5\% & 98.9\% \\
HotpotQA sentence-packed & 100.0\% & 100.0\% & 100.0\% \\
\bottomrule
\end{tabular*}

\vspace{8pt}
\textbf{\textit{(c) Strict matched-state gradient alignment}}\par\vspace{3pt}
\begin{tabular*}{0.82\textwidth}{@{\extracolsep{\fill}}lcc}
\toprule
\textbf{Student context} & \textbf{Answer GA [95\% CI]} & \textbf{Full-evidence GA [95\% CI]} \\
\midrule
Support only & $-0.080$ $[-0.174,0.017]$ & $-0.100$ $[-0.189,-0.009]$ \\
Support + distractors & $-0.053$ $[-0.140,0.035]$ & $-0.025$ $[-0.117,0.075]$ \\
Salience matched & $-0.085$ $[-0.183,0.015]$ & $-0.102$ $[-0.195,-0.005]$ \\
\bottomrule
\end{tabular*}

\vspace{10pt}
\centering
\captionof{table}{Reward-selection and objective controls. Panel (a) applies
PLS as a reward, gate, or weight; panel (b) measures correct-versus-incorrect
rollout separation; panel (c) changes the objective; and panel (d) increases
the update budget. Direct answer supervision remains strongest.}
\label{tab:objective-controls}
\label{tab:pls-control}
\label{tab:sequence-compatibility}
\label{tab:topk-jsd-control}
\label{tab:budget-sweep}
\scriptsize
\setlength{\tabcolsep}{7pt}
\renewcommand{\arraystretch}{1.00}
\textbf{\textit{(a) PLS reward and selection controls (F1, \%)}}\par\vspace{3pt}
\begin{tabular*}{0.90\textwidth}{@{\extracolsep{\fill}}lcccc}
\toprule
\textbf{Intervention} & \textbf{HotpotQA} & \textbf{2Wiki} & \textbf{MuSiQue} & \textbf{QASC} \\
\midrule
Full-evidence baseline & 37.91 & 36.13 & 32.40 & 7.60 \\
PLS as reward & 25.64 & 26.12 & 26.57 & 1.71 \\
PLS gate & \textbf{52.51} & 45.69 & 33.12 & 8.65 \\
PLS weight & 45.59 & \textbf{50.01} & \textbf{34.32} & \textbf{8.78} \\
\bottomrule
\end{tabular*}

\vspace{5pt}
\textbf{\textit{(b) Correct-versus-incorrect rollout separation}}\par\vspace{3pt}
\begin{tabular*}{0.92\textwidth}{@{\extracolsep{\fill}}llcc}
\toprule
\textbf{Dataset (valid)} & \textbf{Teacher} & \textbf{$A_{\mathrm{corr}}/A_{\mathrm{wrong}}$} & \textbf{Difference [95\% CI]} \\
\midrule
\multirow{2}{*}{2Wiki (4,121)}
& Answer & \textminus0.292 / \textminus0.519 & 0.227 [0.213, 0.241] \\
& Full evidence & \textminus0.661 / \textminus1.259 & 0.598 [0.559, 0.637] \\
\midrule
\multirow{2}{*}{MuSiQue (1,235)}
& Answer & \textminus0.273 / \textminus0.657 & 0.384 [0.361, 0.407] \\
& Full evidence & \textminus0.660 / \textminus1.308 & 0.648 [0.563, 0.730] \\
\midrule
\multirow{2}{*}{QASC (891)}
& Answer & \textminus0.152 / \textminus0.209 & 0.057 [0.013, 0.101] \\
& Full evidence & \textminus0.272 / \textminus0.260 & \textminus0.012 [\textminus0.083, 0.051] \\
\bottomrule
\end{tabular*}

\vspace{5pt}
\textbf{\textit{(c) Teacher-top-$20$ JSD objective on HotpotQA}}\par\vspace{3pt}
\begin{tabular*}{0.72\textwidth}{@{\extracolsep{\fill}}lcc}
\toprule
\textbf{Training signal} & \textbf{F1 (\%)} & \textbf{EM (\%)} \\
\midrule
Answer-privileged JSD & 43.1 & 25.0 \\
Full-evidence JSD & \textbf{46.2} & 20.4 \\
Visible-evidence JSD & 45.5 & 19.2 \\
Direct answer supervision & 74.8 & \textbf{60.6} \\
\bottomrule
\end{tabular*}

\vspace{5pt}
\textbf{\textit{(d) HotpotQA update-budget sweep (F1 / EM, \%)}}\par\vspace{3pt}
\begin{tabular*}{0.82\textwidth}{@{\extracolsep{\fill}}lcc}
\toprule
\textbf{Training signal} & \textbf{400 updates} & \textbf{800 updates} \\
\midrule
Direct answer supervision & \textbf{73.16} / \textbf{59.32} & \textbf{74.46} / \textbf{60.99} \\
Answer-only OPD & 47.28 / 31.80 & 53.27 / 38.51 \\
Full-evidence OPD & 46.70 / 26.13 & 51.64 / 32.49 \\
Visible-evidence OPD & 46.64 / 26.10 & 51.46 / 32.25 \\
\bottomrule
\end{tabular*}
\end{table*}

\end{document}